\documentclass[final,5p,times,twocolumn,authoryear]{elsarticle}

\usepackage{amssymb}

\usepackage{float} % auther_import
\usepackage{natbib}
\usepackage[colorlinks]{hyperref} % auther_import
\usepackage{tikz}% auther_import
\usepackage{amsthm} % auther_import
\theoremstyle{definition} % auther_import
\newtheorem{definition}{Definition}
\usetikzlibrary{mindmap}% auther_import
\usepackage{booktabs} %% auther_import
\usepackage{amsmath}  %% auther_import
\biboptions{sort&compress} %% auther_import
\usepackage{array} %% auther_import
\usepackage[figuresright]{rotating} %% auther_import
\usepackage{threeparttable} %% auther_import
\usepackage{framed,multirow} %% auther_import

\usepackage{xcolor} %% auther_import
\usepackage{ulem} %% auther_import
\usepackage{enumitem} %% auther_import

\journal{Neural Networks}

\begin{document}

\begin{frontmatter}


% use the tnoteref command within \title for footnotes;
% use the tnotetext command for theassociated footnote;
% use the fnref command within \author or \affiliation for footnotes;
% use the fntext command for theassociated footnote;
% use the corref command within \author for corresponding author footnotes;
% use the cortext command for theassociated footnote;
% \author{Name\corref{cor1}\fnref{label2}}
% \affiliation{organization={},

\title{A survey on few-shot class-incremental learning}

\author[label1,label2,label4]{Songsong Tian}
\ead{tiansongsong@semi.ac.cn}

\author[label5]{Lusi Li}
\ead{lusili@cs.odu.edu}

\author[label1,label3,label4]{Weijun Li}
\ead{wjli@semi.ac.cn}

\author[label1,label4]{Hang Ran}
\ead{ranhang@semi.ac.cn}

\author[label1,label3,label4]{Xin Ning \corref{cor1}}
\ead{ningxin@semi.ac.cn}
\cortext[cor1]{Corresponding author}

\author[label6]{Prayag Tiwari \corref{cor1}}
\ead{prayag.tiwari@hh.se}
 
\address[label1]{Institute of Semiconductors, Chinese Academy of Sciences, Beijing 100083, China}
\address[label2]{School of Electronic, Electrical and Communication Engineering, University of Chinese Academy of Sciences, Beijing 100049, China}
\address[label3]{School of Integrated Circuits, University of Chinese Academy of Sciences, Beijing 100083, China}
\address[label4]{Beijing Key Laboratory of Semiconductor Neural Network Intelligent Sensing and Computing Technology, Beijing 100083, China}
\address[label5]{Department of Computer Science, Old Dominion University, Norfolk, VA 23529, USA}
\address[label6]{School of Information Technology, Halmstad University, Halmstad 30118, Sweden}

\begin{abstract}
Large deep learning models are impressive, but they struggle when real-time data is not available. Few-shot class-incremental learning (FSCIL) poses a significant challenge for deep neural networks to learn new tasks from just a few labeled samples without forgetting the previously learned ones. 
This setup can easily leads to catastrophic forgetting and overfitting problems, severely affecting model performance. Studying FSCIL helps overcome deep learning model limitations on data volume and acquisition time, while improving practicality and adaptability of machine learning models. This paper provides a comprehensive survey on FSCIL. Unlike previous surveys, we aim to synthesize few-shot learning and incremental learning, focusing on introducing FSCIL from two perspectives, while reviewing over 30 theoretical research studies and more than 20 applied research studies. From the theoretical perspective, we provide a novel categorization approach that divides the field into five subcategories, including traditional machine learning methods, meta learning-based methods, feature and feature space-based methods, replay-based methods, and dynamic network structure-based methods. We also evaluate the performance of recent theoretical research on benchmark datasets of FSCIL. From the application perspective, FSCIL has achieved impressive achievements in various fields of computer vision such as image classification, object detection, and image segmentation, as well as in natural language processing and graph. We summarize the important applications. Finally, we point out potential future research directions, including applications, problem setups, and theory development. Overall, this paper offers a comprehensive analysis of the latest advances in FSCIL from a methodological, performance, and application perspective.

\end{abstract}


\end{frontmatter}


\section{Introduction}
In recent years, significant advancements in computing technology and the widespread availability of large-scale datasets have enabled deep neural networks (DNNs) to make remarkable progresses in various computer vision tasks~\citep{he2016deep,krizhevsky2017imagenet}. However, many of these successes rely on idealized assumptions and massive amounts of available training data, which may not accurately reflect the real-world scenarios where high-quality data is often scarce. For instance, in scenarios where data arrives incrementally in batches and newly added categories contain very few samples, many existing methods prove to be ineffective.

The goal of few-shot class-incremental learning (FSCIL) is to endow AI with the capability to address the aforementioned challenges. This requires DNN models to learn new tasks incrementally from a small number of labeled samples, without forgetting the previously learned ones~\citep{taoFewShotClassIncrementalLearning2020a}. Since Tao first proposed the concept of FSCIL in ~\citet{taoFewShotClassIncrementalLearning2020a}, many scholars have extended it to various application scenarios beyond visual tasks because it conforms to human learning patterns and is suitable for real-world applications.

\begin{figure}%
        \centering
        \includegraphics[width=0.48\textwidth]{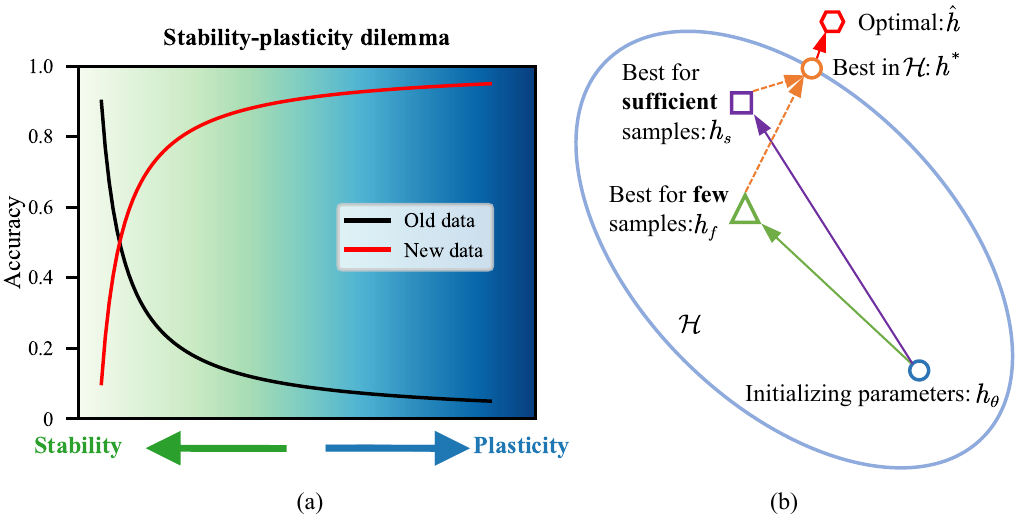}
        \caption{(a) Stability and plasticity cannot be achieved simultaneously. When a model has high stability, it performs well on old data but struggles with new data. As plasticity increases, the model demonstrates enhanced generalization on new data while gradually forgetting old data; (b) Given a hypothesis space $\mathcal{H}$ and initial parameters $h_{\theta}$, $\hat{h}$ is the function that minimizes the expected risk, $h^*$ is the function in $\mathcal{H}$ that minimizes the expected risk. $h_f$ and $h_s$ correspond to the functions of minimizing the empirical risk when data samples are few and sufficient, respectively. When the data is sufficient, ERM yields results closer to $h^*$.}\label{fig-core-challenge}
    \end{figure}

An intuitive method for FSCIL is to fine-tune a base model on a new training set. However, it would lead to catastrophic forgetting~\citep{mccloskeyCatastrophicInterferenceConnectionist1989} and overfitting, corresponding to two core challenges: the stability-plasticity dilemma and unreliable empirical risk minimization.
\begin{itemize}[leftmargin=1em,label={}]
\item \textbf{Stability-plasticity dilemma}

The stability-plasticity dilemma reflects the contradiction between stability and plasticity. Stability means that a neural network should maintain its learned knowledge and resist changes caused by new inputs. Conversely, plasticity means that the network should have the ability to adapt to new inputs or tasks. Catastrophic forgetting can be seen as a manifestation of the stability-plasticity dilemma. In incremental learning (IL), an overly stable model might fail to learn new tasks or data effectively. In contrast, an exceedingly plastic model might rapidly lose information about previously learned tasks or data. See Fig. \ref{fig-core-challenge} (a) for more details.

\item \textbf{Unreliable empirical risk minimization}

In traditional machine learning frameworks, empirical risk minimization (ERM) aims to optimize the average loss on training data. This strategy works well in large-scale data environments where there are enough samples to ensure statistical consistency during training. However, in the context of few shot learning (FSL), this strategy faces a challenge known as the unreliable empirical risk minimizer problem~\citep{wang2020generalizing}.
The core of this problem lies in the fact that when the number of training samples is limited or when there is noise in the samples, the ERM strategy may lead to overfitting. Overfitting means that the model performs well on the training data but has poor generalization performance on new, unseen data. 
This shortfall arises because limited data may not fully represent the true distribution of the entire data generation process, causing the model to capture random noise in the data rather than the underlying true patterns. Fig. \ref{fig-core-challenge} (b) shows that when training samples are insufficient, the ERM function cannot accurately approximate the optimal expected risk minimization function.
\end{itemize}

FSCIL, needs to overcome these two challenges, is even more difficult.
In addition to the challenges mentioned above, due to the large difference in the number of samples between old and new categories, the model tends to bias towards the larger set of old-class training samples during training or prediction, and the imbalance between base and novel class samples also makes it difficult for the model to learn new categories~\citep{Hou_2019_CVPR,taoFewShotClassIncrementalLearning2020a,chenINCREMENTALFEWSHOTLEARNING2021}.
    
Although FSCIL has great potential in real-world applications and has gained significant attentions from researchers, it remains a relatively underexplored area, with a lack of comprehensive reviews. Existing reviews primarily focus on either FSL or IL separately, rather than their combination in FSCIL. 
For example,~\citet{parisi2019continual} focus on continual lifelong learning, though much of the content may not reflect recent advancements. ~\citet{wang2020generalizing} introduced the theoretical foundation of FSL and classified FSL methods from different perspectives.~\citet{belouadah2021comprehensive} provide a summary of Class-IL in visual tasks only.~\citet{zhou2023deep} summarized the latest progress in deep Class-IL from three aspects: data, model, and algorithm.

Our contributions to the field of FSCIL can be summarized as follows:
\begin{itemize}
\item[1)] We conducted an in-depth analysis of fundamental and applied research of FSCIL. Our comprehensive review explores various FSCIL approaches, highlighting their advantages, limitations, and performance on benchmark datasets.
\item[2)] We revisited the theoretical foundations and practical implementations of various FSCIL approaches and proposed a taxonomy of methods based on the underlying approach or technique. This framework provides a useful guide for researchers and practitioners working on FSCIL.
\item[3)] We evaluated the performance of various FSCIL approaches on benchmark datasets, providing insights into the strengths and weaknesses of different methods.
\item[4)] We discussed the potential applications of FSCIL in various domains, such as computer vision, natural language processing, and graph analysis. This analysis highlights the broad range of applications for FSCIL and its potential impact on these fields.
\item[5)] We identified open research challenges and opportunities for future work in the field of FSCIL. This provides a roadmap for future research in the area and helps to guide the direction of future work.
\end{itemize}

The remainder of this paper is organized as follows. Section \ref{sec-2} introduces the problem definition of FSCIL and the relevant research background. 
Section \ref{sec-3} reviews the approaches and notable architectures used in FSL.
Section \ref{sec-4} summarizes the existing FSCIL approaches, including traditional machine learning methods, meta learning-based methods, feature and feature space-based methods, replay-based methods, and dynamic network structure-based methods. Section \ref{sec-5} presents the performance of different FSCIL approaches on benchmark datasets. Section \ref{sec-6} discusses the applications of FSCIL in different domains. Section \ref{sec-7} outlines the future research directions in the FSCIL field. Finally, section \ref{sec-8} concludes the paper.

\section{Problem definition} \label{sec-2}
In supervised learning, we want to learn a function $f\in \mathcal{F}:\mathcal{X}\rightarrow \mathcal{Y}$ that is able to predict the target vector $y\in \mathcal{Y}$, for a given input sample $ x \in \mathcal{X}$. 
To do so, a model is fed with the training data with sufficient instances: $D=\left\{\left(x_i,y_i\right)\right\} _{i=1}^{N}$, which contains independent and identically distributed samples from the distribution $P\left(\mathcal{X},\mathcal{Y}\right)$. 
$x_i\in \mathbb{R}^n$ is a training instance from class $y_i\in \mathcal{Y}$ and $\mathcal{Y}$ is the corresponding label space. In order to train this function $f$, we minimize the expected risk over the instance distribution $P$:
\begin{equation} \label{eq:expected risk}
\varepsilon _{ex}=\mathbb{E}_{\left( x,y \right) \sim P\left( \mathcal{X},\mathcal{Y} \right)}\left[ \ell \left( f\left( x \right) ,y \right) \right], 
\end{equation}
where $\ell(\cdot,\cdot )$ captures the discrepancy between prediction and ground-truth label. However, the joint distribution  $P$ in unknown, therefore the learning algorithm actually aims at minimizing the empirical risk: 
\begin{equation} \label{eq:empirical risk}
\varepsilon _{em}=\mathbb{E}_{\left( x,y \right) \sim D}\left[ \ell \left( f\left( x \right) ,y \right) \right], 
\end{equation}
\subsection{Problem formalization}\label{sec:Problem formalization}
    \begin{figure}%
        \centering
        \includegraphics[]{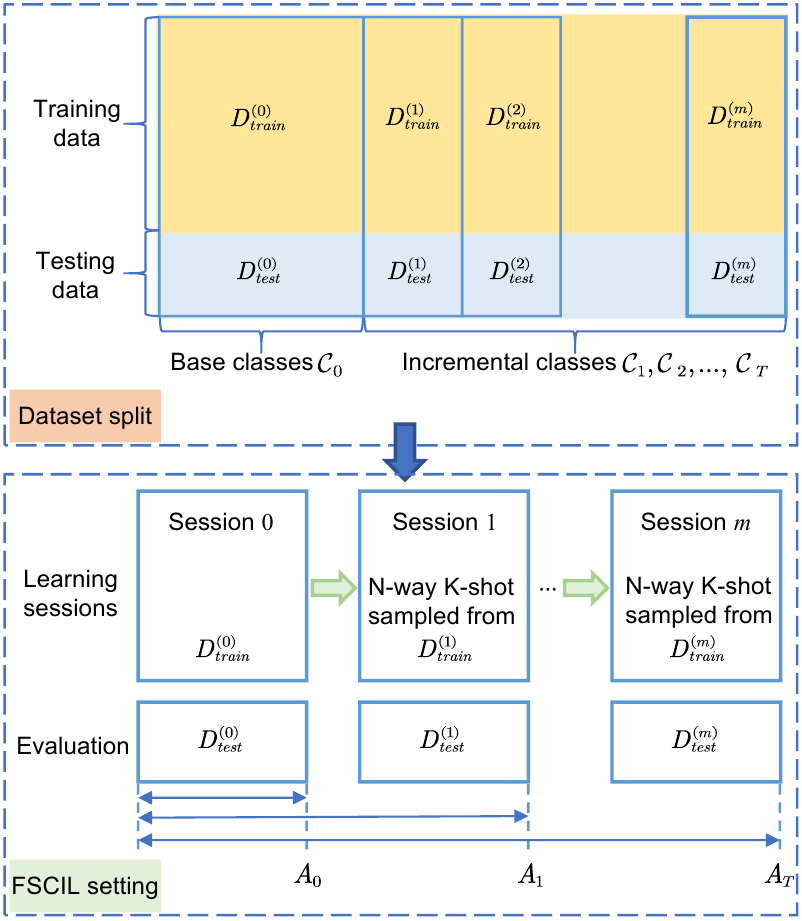}
        \caption{Dataset setting. Figure adapted from~\citet{zhengFewShotClassIncrementalLearning2021}}\label{dataset_setting}
\end{figure}
Fig. \ref{dataset_setting} shows the form of dataset split and the way of FSCIL experiment setup. FSCIL task comprises a base session with sufficient training data and multiple incremental sessions with limited training data. 
The learning process within each session involves only the data relevant to the current task, while the model is also required to preserve the knowledge of previous tasks when acquiring new ones.
The task is to train the model from a continuous data stream in a class-incremental form.

The FSCIL problem is defined as follows. Here we assume an $m$-step FSCIL task. Let $\left\{ D_{train}^{\left( 0 \right)},D_{train}^{\left( 1 \right)},...,D_{train}^{\left( m \right)} \right\} 
$ and $\left\{ D_{test}^{\left( 0 \right)},D_{test}^{\left( 1 \right)},...,D_{test}^{\left( m \right)} \right\} 
$ denote the training and testing data for sessions $\left\{ 0,1,...,m \right\}$, respectively. 
For session $j$, it has training data $D_{train}^{\left( j \right)}$ with the corresponding label space of $\mathcal{Y}_j$. Training data from different sessions are disjoint, that is, $\mathcal{Y}_a\cap \mathcal{Y}_b=\varnothing \left( a\ne b \right)$. The limited instances in $D_{train}^{\left( j \right)}$ can be organized as $N$-way $K$-shot data format, i.e., there are $N$ classes in the dataset, and each class has $K$ training images.
Facing a new dataset $D_{train}^{\left( j \right)}$, a model should learn new classes and meanwhile maintain performance over old classes, i.e., minimize the expected risk $R$ over all the seen classes:
\begin{equation} \label{eq:FSCIL}
\mathbb{E}_{\left( x,y \right) \sim D_{train}^{\left( 0 \right)}\cup D_{train}^{\left( 1 \right)}\cup ...\cup D_{train}^{\left( j \right)}}\left[ \ell \left( f\left( x;D_{train}^{\left( j \right)},W^{j-1} \right) ,y \right) \right] 
, 
\end{equation}
In Eq. \ref{eq:FSCIL}, the learning algorithm $f$ should build the new model
based on new dataset $D_{train}^{\left( j \right)}$ and current old model $W^{j-1}$, and
minimize the loss over all seen classes.
During testing, the model will be evaluated on all seen classes so far. For session $j$, its testing data $D_{test}^{\left( j \right)}$ has the corresponding label space of $\mathcal{Y}_0\cup \mathcal{Y}_1...\cup \mathcal{Y}_i$.

\subsection{Relevant learning problems}
\textbf{Few-shot Learning.} Humans are very skilled at identifying a new object with very few samples. For example, a child can recognize what a "zebra" or "rhinoceros" is with just a few pictures from a book. Inspired by human's rapid learning ability, researchers hope that machine learning models can quickly learn new categories with only a small number of samples after learning a large amount of data for a certain number of categories. This is the problem that FSL aims to solve. In recent years, the concept of FSL has received widespread attention, and there have been many outstanding algorithm models in the field of image classification~\citep{snell2017prototypical, zhang2018deep, finn2017model}. There are mainly three categories of FSL methods: fine-tune based, data augmentation based, and transfer learning based. 

Considering a learning task $T$, FSL deals with a data set $D=\left\{ D_{train},D_{test} \right\}$. It consists of a training set $D_{train}=\left\{ \left( x_i,y_i \right) \right\} _{i=1}^{I}$, where $I$ is small, and a testing set $D_{test}=\left\{ x_{test} \right\}$. Usually, one considers the $N$-way $K$-shot classification in which $D_{train}$ contains $I=KN$ examples from $N$ classes each with $K$ examples. FSL is mainly a supervised learning problem~\citep{wang2020generalizing}. Due to the small size of $D_{train}$, the model bias, $\varepsilon =\left| \varepsilon _{ex}-\varepsilon _{em} \right|$, is too large, making it hard to learn a high-quality prediction function $f\in \mathcal{F}:\mathcal{X}\rightarrow \mathcal{Y}$.
    
\textbf{One-shot Learning.} In the late 1980s and 1990s, some researchers already noticed the problem of one-shot learning. It was not until 2003 that~\citet{fe2003bayesian} formally introduced the concept. They believed that when there is only one or a few labeled samples for a new category, the previously learned old categories can help predict the new category~\citep{fei2006one}. In the $N$-way $K$-shot paradigm, when $N=1$, FSL is called one-shot learning problem. Since the settings are similar, it is not necessary to distinguish between the two concepts in most cases.

\textbf{Zero-shot Learning.} In the $N$-way $K$-shot paradigm, FSL becomes a zero-shot learning problem (ZSL) when $N=0$. ZSL was first introduced by~\citet{palatucci2009zero}. 
Since ZSL does not contain examples with supervised information, it recognizes new sample categories by utilizing semantic label attribute information in the absence of training samples. This approach is inspired by human learning and reasoning capabilities, allowing computers to possess transfer and reasoning abilities. Specifically, a training data for ZSL is formulated as $S=\left\{ \left( x,y,a\left( y \right) \right) |x\in \mathcal{X}_S,y\in \mathcal{Y}_S,a\left( y \right) \in \mathcal{A} \right\}$, where $\mathcal{X}_S$ is set of image/features from seen classes, $\mathcal{Y}_S$ is set of seen class labels, $a(y)$ is semantic embedding for class $y$. The test set  is formulated as $U=\left\{ \left( x,y,a\left( y \right) \right) |x\in \mathcal{X}_U,y\in \mathcal{Y}_U,a\left( y \right) \in \mathcal{A} \right\}$,  where $\mathcal{X}_U$ is set of unseen class image/features, $\mathcal{Y}_U$ is set of unseen class labels, $\mathcal{Y}_U\cap \mathcal{Y}_C=\varnothing$.
    
\textbf{Meta Learning.} Meta learning is often understood as learning to learn. It is the process of extracting the experience of multiple learning episodes and using this experience to improve future learning performance~\citep{hospedales2022Meta}. Meta learning is usually divided into two stages. In the meta-training stage, the model is trained using multiple source (or training) tasks to obtain initial network parameters with strong generalization ability. In the meta-testing stage, the settings of the new tasks are the same as those of the source tasks, but these samples have not been seen during the training process. Each task in the training tasks or testing tasks is divided into a support set and a query set. Meta learning has wide applications in the fields of computer vision, reinforcement learning, and architecture search. Meta learning is naturally suitable for FSL, and many studies have used meta-learning as a means of FSL, enabling the model to learn from a small number of new task samples~\citep{ren2018meta, elsken2020meta, jamal2019task}.

\textbf{Transfer Learning.} Transfer Learning~\citep{zhuang2020comprehensive} focuses on the transfer of knowledge across different domains, enabling the transfer of knowledge from domains/tasks with abundant training data to novel domains/tasks with scarce training data. Its definition is as follows.

\begin{definition}
\label{def-transfer_learning}
\textbf{Transfer learning}. \textit{Given a source domain $D_S$ and a corresponding task $T_S$, a target domain $D_T$ and a corresponding task $T_T$. The primary aim of transfer learning is to leverage the knowledge obtained from $D_S$ and $T_S$ to enhance the learning performance of $D_T$ and $T_T$, where $D_S \neq D_T$ or $T_S \neq T_T$}~\citep{pan2010survey}.
\end{definition}

The key to successful knowledge transfer is the presence of a connection between the two learning activities. If there are few commonalities between domains, knowledge transfer may fail and have a negative impact on the new task. In everyday life, people engage in many instances of transfer learning, such as learning to ride a bike, which makes it easier to learn how to ride a motorcycle. Transfer learning can reduce the reliance on large amounts of target domain data when constructing learning machines. As a result, it has broad applications in zero-shot and few-shot domains, including style transfer, feature space transfer for data augmentation, and label-efficient learning of transferable representations across domains~\citep{azadi2018multi,liu2018feature,luo2017label}.
    
\textbf{Incremental Learning.} The definition of IL can also be expressed using Eq. \ref{eq:FSCIL}, but the difference from FSCIL is that there are plenty of samples for each incremental category. IL is also known as continuous learning, lifelong learning, or never-ending learning, is a field of machine learning that is gaining increasing attention. It is typically used to address the problem of catastrophic forgetting, where performance on previously learned tasks deteriorates sharply after learning new tasks. The ability of IL is to continuously process a stream of information from the real world while retaining, integrating, and optimizing old knowledge at the same time. The methods proposed in IL are broadly categorized into three categories: replay-based methods, regularization-based methods, and parameter isolation methods~\citep{de2021continual}.~\citet{van2019three} proposed three scenarios for IL, including Task-IL, Domain-IL, and Class-IL. And Class-IL is considered the most difficult one since the newly added classes often exhibit high similarity with the already learned classes. Currently, only replay-based methods produce acceptable results for Class-IL.

\subsection{Variants of few-shot class incremental learning}
\textbf{Generalized few-shot incremental learning.}
Before the emergence of FSCIL, similar settings had been proposed in previous research, such as those presented by~\citep{qi2018low,gidaris2018dynamic,yoonXtarNetLearningExtract2020,xie2019meta}. These studies introduced Generalized Few-Shot Incremental Learning (GFSIL). Specifically, a pre-trained model will learn new classes with limited instances. The goal of GFSIL is to maintain classification performance for both old and new classes. However, GFSIL only has one incremental phase, and its data partitioning format is different from FSCIL. For example, CIFAR-100 is randomly divided into 40, 10, and 50 categories, which serve as the meta-training, meta-validation, and meta-testing sets respectively. GFSIL is considered less challenging than FSCIL.
To address the challenge of GFSIL,~\citet{qi2018low} proposes a solution that utilizes the average feature initialization method with few shots to initialize new class representations. Meanwhile,~\citet{gidaris2018dynamic} introduces dynamic few-shot learning to avoid forgetting, which employs a novel attention-based weight generator for few-shot classification. The dot-product calculation method is replaced with the cosine-similarity function to incorporate the few-shot classification weight generator into the recognition system.~\citet{renIncrementalFewShotLearning2019} proposes an Attention Attractor Network to regulate the learning of novel classes. Additionally,~\citet{yoonXtarNetLearningExtract2020} suggests a method for fusing base features, while~\citet{ye2021learning} puts forward the idea of synthesizing few-shot classifiers with a shared neural dictionary.~\citet{xie2019meta} introduces Meta Module Generation (MetaMG) which utilizes meta-learning to learn a set of meta-modules, which are small neural networks that can be quickly adapted to new tasks. During the IL process, the MetaMG approach uses the learned meta-modules to generate task-specific modules for new classes.

\textbf{Few-shot incremental learning.}
Similar to FSCIL,~\citet{ayub2020cognitively} examines the problem of few-shot incremental learning (FSIL) and proposes a cognitively-inspired approach. They represent each image class as a centroid. In the experimental setting of FSIL, the number of classes for both base and incremental is the same, which differs from the rich base data setting in FSCIL. Additionally, in order to tackle the issue of the inability to learn from data streams in ZSL,~\citet{wei2020lifelong,weiIncrementalZeroShotLearning2021} have proposed the concept of incremental zero-shot learning (IZSL). Unlike traditional ZSL, IZSL involves multiple learning phases for new classes.

\textbf{Incremental few-shot object detection.}
In the setting of incremental few-shot object detection (iFSD)~\citet{perez2020incremental}, abundant base-class samples and a few novel-class samples are available. The model can use all the base-class samples for bootstrapping as prior knowledge is required for the model to learn in the few-shot way. Equipped with the prior knowledge of base-class data, the model cannot visit base-class samples again when learning knowledge of novel classes. In other words, the model with prior knowledge should be able to learn from the few samples of unseen categories without relearning basic knowledge, which is aligned with the practical application scenes where the pre-trained model should be competent to adapt to unseen information incrementally.

Despite many studies sharing similar settings to FSCIL, the current mainstream in academia still focuses on FSCIL. Therefore, this review primarily focuses on the more challenging FSCIL research. 

% color schedule:Few-Shot Class-Incremental Learning via Class-Aware Bilateral Distillation

\section{Methods for few-shot learning} \label{sec-3}
For FSL tasks, specialized network architectures or tricks are typically required to handle limited annotated data. In FSCIL research, many methods build upon advancements in FSL. In this section, we focus on providing a brief overview of commonly used network architectures in FSL, without discussing the novelty or effectiveness of the methods. And they might not represent the latest research.

Numerous surveys have been conducted on the topic of FSL, proposing various classification approaches~\citep{wang2020generalizing, song2023comprehensive, jadon2020overview}. One straightforward approach is to categorize FSL into four categories: data augmentation methods, metric-based methods, model-based methods, and optimization-based methods~\citep{jadon2020overview}. Hereafter, we will provide a brief introduction to the commonly used network architectures within these four categories.

\subsection{Data augmentation methods}
In FSL, data augmentation is an important strategy. It alleviates the problem of data scarcity by increasing the diversity of existing data, rather than collecting new data. Data augmentation significantly reduces the risk of overfitting and effectively enhances the model's generalization ability. 
Data augmentation can be categorized by its source: transforming samples from the training set, transforming samples from a weakly labeled or unlabeled data set, or transforming samples from similar data sets~\citep{wang2020generalizing}.
Besides directly augmenting the data, one can also train a model to generate new samples or features~\citep{kong2022few}, such as VAEs or GANs, to achieve the goal of data augmentation.

\subsection{Metric-based methods}
\begin{figure*}[t]%
        \centering
        \includegraphics[]{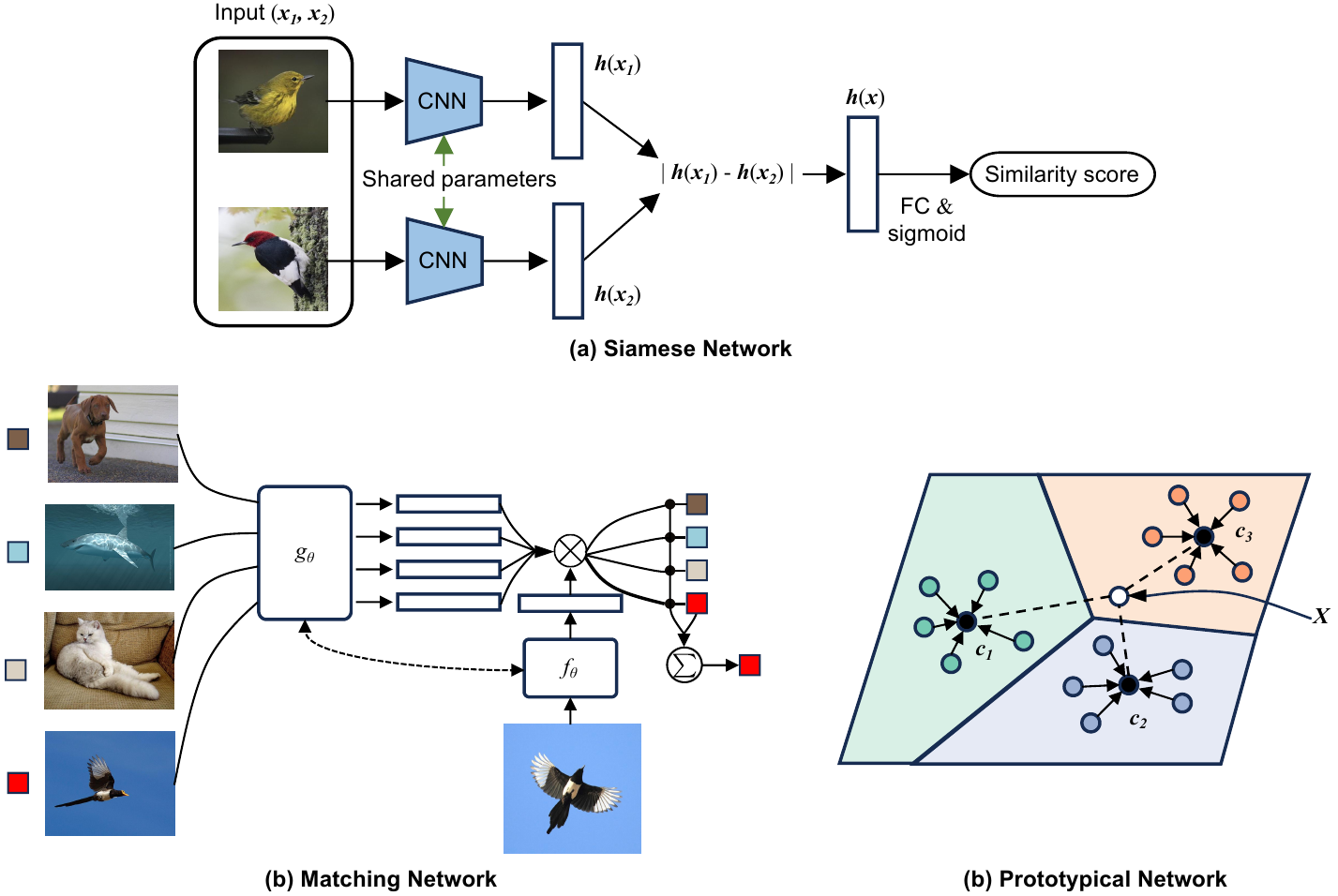}
        \caption{Common network architectures in metric-based methods: \textbf{(a) Siamese Networks}: Utilize twin subnetworks to extract features from two input samples and compute the distance between these features;
\textbf{(b) Matching Networks}: By using the attention mechanism to dynamically match and aggregate the support set and query set examples, Matching Networks can generate class-related feature representations for query samples; \textbf{(c) Prototypical Networks}: Represent each class by the mean of their features. Thus, in the embedding space, closer features are more likely to belong to the same class. \label{fig-fsl}}
\end{figure*}

Methods based on metrics classify objects in the embedded space by computing the similarity or distance between samples in the support set and the query set. For instance, by calculating the Euclidean distance between a test sample and each class in the support set, the test sample is assigned to the category of the nearest support set sample. In FSL, commonly used metric learning methods include Siamese Network~\citep{koch2015siamese}, Matching Network~\citep{vinyals2016matching}, and Prototypical Network~\citep{snell2017prototypical}. Fig. \ref{fig-fsl} illustrates the network structure differences among these three methods. These methods do not require extensive data but optimize metrics to ensure similar samples are close, while samples of different classes are distant.

\subsection{Model-based methods}
Model-based methods primarily refer to designing or using specific network architectures to address FSL challenges. For instance, Memory-Augmented Neural Networks (MANN)~\citep{santoro2016meta} use external memory spaces to explicitly store class information, thus leveraging the long-term memory capabilities inherent in neural networks for FSL tasks. Meta Networks~\citep{munkhdalai2017meta} learn meta-level knowledge across tasks and adjust their inductive biases through quick parameterization for swift generalization. These network structures efficiently utilize a limited number of labeled samples for rapid learning and adaptation.
\subsection{Optimization-based methods}
Optimization-based methods focus on adjusting the training strategy of models to adapt to situations with limited annotated data. It typically involves modifying the loss function, regularization terms, or the optimization algorithm itself to ensure that the models can quickly converge on few-shot data without overfitting. 
For example, Model-Agnostic Meta-Learning (MAML)~\citep{finn2017model} is a common optimization technique that quickly learns knowledge from limited new data. It trains the model's initial parameters using various datasets to ensure peak performance when tackling new tasks. Building on MAML, Reptile~\citep{nichol2018reptile} simplifies computational complexity by reducing gradient calculations from two steps to one, thereby increasing computational speed.

\section{Few-shot class-incremental learning: taxonomy} \label{sec-4}
% FSCIL approaches
 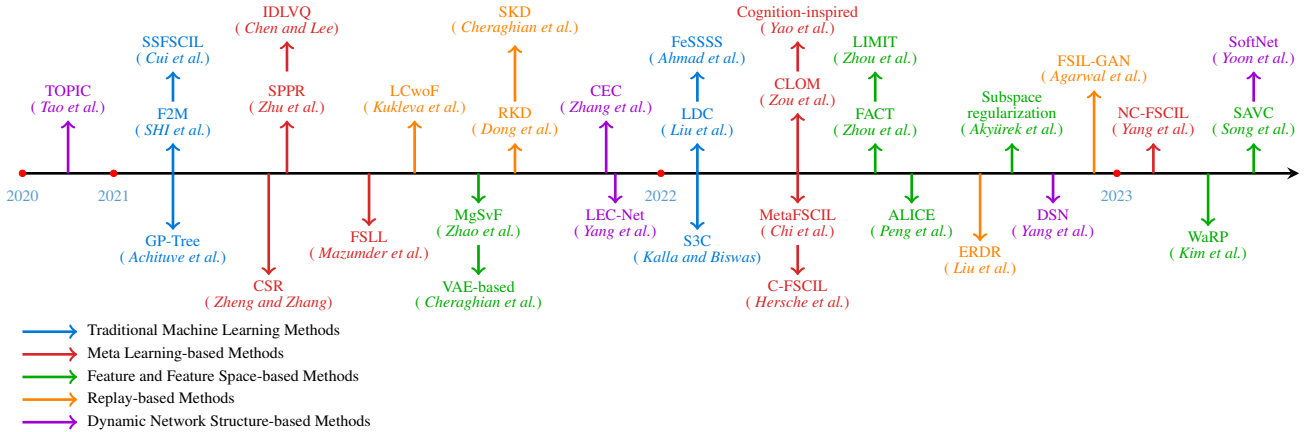
\begin{figure*}[h]
\centering
\begin{tikzpicture}[x=1.2cm,y=1cm,every node/.style = {font=\fontsize{6}{6}\selectfont}]
\hypersetup{hidelinks}
% \definecolor{traditional}{RGB}{233, 29, 39}
% \definecolor{meta}{RGB}{54, 125, 183}
% \definecolor{feature}{RGB}{74, 173, 74}
% \definecolor{replay}{RGB}{231, 126, 31}
% \definecolor{dynamic}{RGB}{117, 112, 179}

% \definecolor{traditional}{RGB}{58, 129, 229}
% \definecolor{meta}{RGB}{230, 92, 78}
% \definecolor{feature}{RGB}{76, 174, 80}
% \definecolor{replay}{RGB}{248, 152, 56}
% \definecolor{dynamic}{RGB}{153, 80, 207}

\definecolor{traditional}{RGB}{0, 120, 215}
\definecolor{meta}{RGB}{215, 40, 40}
\definecolor{feature}{RGB}{0, 170, 0}
\definecolor{replay}{RGB}{255, 130, 0}
\definecolor{dynamic}{RGB}{160, 0, 210}

\definecolor{lightblue}{RGB}{91,155,212}

% draw x-axis
% \draw[-stealth, pink,line width=1.8pt] (-1,0) -- (11,0) node[right] {$x$};
\draw[-stealth, black,line width=1.pt] (-1,0) -- (13,0) node[right] {};

\foreach \x in {-1,0,6,11}
\filldraw[red] (\x,0) circle (1.2pt) node[below] {};

% Draw labels
\node[below] at (-1,-0.1) {\textcolor{lightblue}{2020}};
\node[below] at (0,-0.1) {\textcolor{lightblue}{2021}};
\node[below] at (6,-0.1) {\textcolor{lightblue}{2022}};
\node[below] at (11,-0.1) {\textcolor{lightblue}{2023}};

% Draw timeline events with arrows and legend in pink color
\node[above,dynamic,align=center] at ( -0.5,0.65) {TOPIC\\(\textit{~\citeauthor{taoFewShotClassIncrementalLearning2020a}})};
\draw[->,dynamic,line width=1pt] (-0.5,0) -- (-0.5,0.7);

\node[above,traditional,align=center] at (0.65,0.35) {F2M\\(\textit{~\citeauthor{shiOvercomingCatastrophicForgetting2021}})};
\draw[->,traditional,line width=1pt] (0.65,0) -- (0.65,0.4);

\node[below,traditional,align=center] at ( 0.65,-0.7) {GP-Tree\\(\textit{~\citeauthor{achituveGPTreeGaussianProcess2021}})};
\draw[->,traditional,line width=1pt] (0.65,0) -- (0.65, -0.75);

\node[above,traditional,align=center] at (0.65,1.3) {SSFSCIL\\(\textit{~\citeauthor{cui2021semi}})};
\draw[->,traditional,line width=1pt] (0.65,0.95) -- (0.65,1.35);

\node[above,meta,align=center] at (1.9,0.65) {SPPR\\(\textit{~\citeauthor{zhu2021self}})};
\draw[->,meta,line width=1pt] (1.9, 0) -- (1.9,0.7 );

\node[below,meta,align=center] at (1.7,-1.3) {CSR\\(\textit{~\citeauthor{zhengFewShotClassIncrementalLearning2021}})};
\draw[->,meta,line width=1pt] (1.7, 0) -- (1.7, -1.35);

\node[above,meta,align=center] at (1.9,1.7) {IDLVQ\\(\textit{~\citeauthor{chenINCREMENTALFEWSHOTLEARNING2021}})};
\draw[->,meta,line width=1pt] (1.9,1.35 ) -- (1.9, 1.75);

\node[below,meta,align=center] at (2.8,-0.65) {FSLL\\(\textit{~\citeauthor{mazumderFewShotLifelongLearning2021}})};
\draw[->,meta,line width=1pt] (2.8, 0) -- (2.8, -0.7);

\node[below,feature,align=center] at (4,-1.3) {VAE-based\\(\textit{~\citeauthor{cheraghian2021synthesized}})};
\draw[->,feature,line width=1pt] (4, -0.95) -- (4, -1.35);

\node[below,feature,align=center] at (4,-0.35) {MgSvF\\(\textit{~\citeauthor{zhaoMgSvFMultiGrainedSlow2021}})};
\draw[->,feature,line width=1pt] (4, 0) -- (4, -0.4);

\node[above,replay,align=center] at (3.3,0.65) {LCwoF\\(\textit{~\citeauthor{kuklevaGeneralizedIncrementalFewShot2021}})}; 
\draw[->,replay,line width=1pt] (3.3, 0) -- (3.3,0.7 );

\node[above,replay,align=center] at (4.4,1.7) {SKD\\(\textit{~\citeauthor{cheraghianSemanticAwareKnowledgeDistillation2021}})};
\draw[->,replay,line width=1pt] (4.4,0.95 ) -- (4.4, 1.7);

\node[above,replay,align=center] at (4.4,0.35) {RKD\\(\textit{~\citeauthor{dongFewShotClassIncrementalLearning2021a}})};
\draw[->,replay,line width=1pt] (4.4,0 ) -- (4.4, 0.4);

\node[below,dynamic,align=center] at (5.5,-0.35) {LEC-Net\\(\textit{~\citeauthor{yangLearnableExpansionandCompressionNetwork2021}})};
\draw[->,dynamic,line width=1pt] (5.5, 0) -- (5.5, -0.4);

\node[above,dynamic,align=center] at (5.4,0.65) {CEC\\(\textit{~\citeauthor{zhang2021few}})};
\draw[->,dynamic,line width=1pt] (5.4,0 ) -- (5.4,0.7 );

% scale 6：
\node[above,traditional,align=center] at (6.4,0.35) {LDC\\(\textit{~\citeauthor{liuLearnableDistributionCalibration2022}})};
\draw[->,traditional,line width=1pt] (6.4,0 ) -- (6.4, 0.4);

\node[below,traditional,align=center] at (6.4,-0.7) {S3C\\(\textit{~\citeauthor{kalla2022s3c}})};
\draw[->,traditional,line width=1pt] (6.4,0 ) -- (6.4, -0.75);

\node[above,traditional,align=center] at (6.4,1.3) {FeSSSS\\(\textit{~\citeauthor{ahmadFewShotClassIncremental2022}})};
\draw[->,traditional,line width=1pt] (6.4, 0.95) -- (6.4, 1.35);

\node[below,meta,align=center] at (7.5,-0.35) {MetaFSCIL\\(\textit{~\citeauthor{chiMetaFSCILMetaLearningApproach2022}})};
\draw[->,meta,line width=1pt] (7.5, 0) -- (7.5, -0.4);

\node[above,meta,align=center] at (7.5,0.75) {CLOM\\(\textit{~\citeauthor{zou2022marginbased}})};
\draw[->,meta,line width=1pt] (7.5,0 ) -- (7.5,0.8 );

\node[below,meta,align=center] at (7.5,-1.3) {C-FSCIL\\(\textit{~\citeauthor{herscheConstrainedFewShotClassIncremental2022}})};
\draw[->,meta,line width=1pt] (7.5, -0.95) -- (7.5, -1.35);

\node[above,meta,align=center] at (7.5,1.7) {Cognition-inspired\\(\textit{~\citeauthor{yao2022few}})};
\draw[->,meta,line width=1pt] (7.5,1.35 ) -- (7.5, 1.75);

\node[above,feature,align=center] at (8.35,0.35) {FACT\\(\textit{~\citeauthor{zhouForwardCompatibleFewShot2022a}})};
\draw[->,feature,line width=1pt] (8.35 , 0) -- ( 8.35,0.4);

\node[above,feature,align=center] at ( 8.35,1.3) {LIMIT\\(\textit{~\citeauthor{zhouFewShotClassIncrementalLearning2022}})};
\draw[->,feature,line width=1pt] ( 8.35,0.95 ) -- ( 8.35, 1.35);

\node[below,feature,align=center] at ( 8.75,-0.35) {ALICE\\(\textit{~\citeauthor{pengFewShotClassIncrementalLearning2022}})};
\draw[->,feature,line width=1pt] (8.75 ,0 ) -- (8.75 ,-0.4);

\node[above,feature,align=center] at ( 9.85,0.35) {Subspace\\regularization\\(\textit{~\citeauthor{akyurekSubspaceRegularizersFewShot2022}})};
\draw[->,feature,line width=1pt] ( 9.85, 0) -- (9.85 ,0.4);

\node[below,replay,align=center] at ( 9.5,-0.85) {ERDR\\(\textit{~\citeauthor{liuFewShotClassIncrementalLearning2022}})};
\draw[->,replay,line width=1pt] ( 9.5, 0) -- (9.5 ,-0.9);

\node[above,replay,align=center] at ( 10.75,1.05) {FSIL-GAN\\(\textit{~\citeauthor{agarwal2022semantics}})};
\draw[->,replay,line width=1pt] (10.75 ,0 ) -- (10.75 ,1.1);

\node[below,dynamic,align=center] at (10.3 ,-0.35) {DSN\\(\textit{~\citeauthor{yang2022dynamic}})};
\draw[->,dynamic,line width=1pt] ( 10.3,0 ) -- ( 10.3,-0.4);

% scale 11.
\node[above,meta,align=center] at ( 11.4,0.35) {NC-FSCIL\\(\textit{~\citeauthor{yang2023neural}})};
\draw[->,meta,line width=1pt] ( 11.4, 0) -- (11.4 ,0.4);

\node[below,feature,align=center] at ( 12,-0.65) {WaRP\\(\textit{~\citeauthor{kim2023warping}})};
\draw[->,feature,line width=1pt] ( 12, 0) -- (12 ,-0.7);

\node[above,feature,align=center] at ( 12.5,0.35) {SAVC\\(\textit{~\citeauthor{song2023learning}})};
\draw[->,feature,line width=1pt] ( 12.5, 0) -- (12.5 ,0.4);

\node[above,dynamic,align=center] at ( 12.5,1.3) {SoftNet\\(\textit{~\citeauthor{yoon2023soft}})};
\draw[->,dynamic,line width=1pt] ( 12.5, 0.95) -- (12.5 ,1.35);

% Draw timeline events with arrows and legend in pink color

\draw[->,traditional,line width=1pt] (-1.,-2.1) -- (-0.4,-2.1) node[right,black] {Traditional Machine Learning Methods};
\draw[->,meta,line width=1pt] (-1.,-2.4) -- (-0.4,-2.4) node[right,black] {Meta Learning-based Methods};
\draw[->,feature,line width=1pt] (-1.,-2.7) -- (-0.4,-2.7) node[right,black] {Feature and Feature Space-based Methods};
\draw[->,replay,line width=1pt] (-1.,-3.0) -- (-0.4,-3.0) node[right,black] {Replay-based Methods};
\draw[->,dynamic,line width=1pt] (-1.,-3.3) -- (-0.4,-3.3) node[right,black] {Dynamic Network Structure-based Methods};

\end{tikzpicture}
\caption{Chronological overview of key FSCIL research developments.\label{fig-chronological}}
\end{figure*}
For fundamental research on FSCIL, there is currently no unified classification standard.~\citet{zou2022marginbased} divided FSCIL into metric-based and fine-tuning-based methods.
The metric-based method is similar to the concept of FSL~\citep{snell2017prototypical,vinyals2016matching}, and its key issue lies in the prototype representation and similarity metric. 
In FSCIL, the fine-tuning-based approaches are widely used, and we refer to this method as Base Classes Pretraining and Novel Classes Fine-tuning (BPNF).

\begin{definition}
\label{def-BPNF}
\textbf{Base Classes Pretraining and Novel Classes Fine-tuning (BPNF)} \textit{is a common approach used in FSCIL, which involves pre-training a model on data-rich base data and fine-tuning the model to better fit the novel classes in the incremental phase. This approach leverages the knowledge learned from the base classes to improve the model's performance on novel, unseen classes. }
\end{definition}

However, the above classification method is too broad and not suitable for many FSCIL studies. In this paper, we have summarized 33 advanced studies and categorized them into five families based on the key point or technique used in FSCIL:

\begin{itemize}
    \item Traditional machine learning methods
    \item Meta learning-based methods
    \item Feature and feature space-based methods
    \item Replay-based methods
    \item Dynamic network structure-based methods
\end{itemize}

Fig. \ref{fig-chronological} displays an  approach classification chart for corresponding years chronologically. It is worth noting that although the experimental settings in FSCIL often involve the idea of meta-learning, these methods are not classified as meta learning-based methods because the key points of the methods used are not based on meta-learning techniques.

\subsection{Traditional machine learning methods}
\subsubsection{Supervised learning strategies}
The capacity of a model that has undergone fine-tuning through an incremental process is limited by the amount of new class sample data available. To alleviate this constraint, certain studies have introduced additional semi-supervised or unsupervised data, in addition to relying solely on labeled supervised data, to refine the supervision method.

In~\citet{cui2021semi}, semi-supervised learning was introduced to FSCIL and, based on the setting in~\citet{taoFewShotClassIncrementalLearning2020a}, 50 unlabeled data were introduced in each incremental session. During the training process, the unlabeled data were combined with labeled data to enhance the performance of FSCIL.
In~\citet{ahmadFewShotClassIncremental2022}, leveraging self-supervised learning was proposed to alleviate overfitting and catastrophic forgetting. Specifically, in addition to training the ResNet-18 model with base-class data, a deeper ResNet-50 network was trained using self-supervised methods on a large dataset. These two networks were then frozen to possess two powerful feature extractors. Two sets of feature vectors were input into a Gaussian Generator to learn models for new classes while passing their features. Subsequently, through feature fusion plus classifier, the forgetting can be effectively countered, and adaptation to the emergence of new classes can be achieved.
For the first time,~\citet{kalla2022s3c} proposed the self-supervised stochastic classifier (S3C) to solve FSCIL. The stochasticity of the classifier avoids overfitting to few-shot novel classes, while combining self-supervised training enables better preservation of base-class knowledge.

\subsubsection{Statistical distribution}
From the statistical distribution perspective, solving the FSCIL problem involves fitting models to existing datasets and predicting the data distribution of the classes, which has excellent model interpretability. To address the limitations of common Gaussian process classification in large-scale class classification tasks,~\citet{achituveGPTreeGaussianProcess2021} proposed GP-Tree. GP-Tree is a tree-based hierarchical model that uses Polya-Gamma data augmentation to fit data to a Gaussian process, which can adapt well to the number of classes and data size.~\citet{liuLearnableDistributionCalibration2022} proposed the learnable distribution calibration (LDC) approach, which is rooted in a parameterized calibration unit (PCU). PCU initializes the feature distribution of each class by using a Gaussian sampler defined by the mean vector and stored covariance matrix to generate a set of feature samples. Specifically, the Gaussian sampler generates enough feature samples during IL to form biased distributions for old and new classes. The PCU cyclically updates the generated feature samples, thereby restoring the old class distribution and calibrating the new class distribution. Due to the fixed size of the covariance matrix, this method has low memory consumption. Both methods achieve good results in FSCIL, but the drawback is that the modeling process is complex.

\subsubsection{Function optimization}
Existing methods focus on overcoming catastrophic forgetting when learning new tasks, while~\citet{shiOvercomingCatastrophicForgetting2021} have analyzed this issue from the perspective of function optimization and found that flat local minima obtained during training on base classes have better generalization ability than sharp minima. 
Flat minima is a crucial concept in machine learning and optimization theory~\citep{hochreiter1997flat}. In the vicinity of flat minima, minor parameter alterations do not significantly impact the loss function, leading to models with robustness. Furthermore, flat minima serve as a natural form of regularization, typically preventing models from overfitting and enhancing their generalization capabilities.
Specifically,~\citet{shiOvercomingCatastrophicForgetting2021} suggest searching for flat local minima of the base training objective function and then fine-tune the model parameters within the flat region on new tasks, substantially reducing catastrophic forgetting.

\subsection{Meta learning-based methods}
In the realm of FSL or IL, meta-learning can leverage existing knowledge to address current learning problems, and improve the stability and reliability of the system through continuous knowledge accumulation. In FSL, meta-learning enhances the learning effect of the current task by utilizing data from other related tasks~\citep{finn2017model, rusu2018metalearning, snell2017prototypical, liu2020prototype}. In IL, meta-learning can be used to reduce dependence on new data, thereby avoiding overfitting~\citep{riemer2018learning}. It is natural to apply meta-learning to FSCIL.

Here, we divide the meta learning-based FSCIL method into two categories: prototype learning-based, and meta process-based method.

\subsubsection{Prototype learning}
Prototype learning aims to identify a small set of exemplars that accurately represent a given dataset, and then use the similarity between the data points and the prototypes to classify new data points or complete other visual tasks. Commonly used class prototypes are defined as follows:
\begin{equation} \label{eq:prototype}
\boldsymbol{\mu }_c=\frac{1}{\left| S_c \right|}\sum_{\boldsymbol{x}\in S_c}{f_{\theta}\left( \boldsymbol{x} \right)},
\end{equation}
where $S_c$ is the set of all samples from class $c$; $f_{\boldsymbol{\theta }}$ is the embedding network parameterized by $\boldsymbol{\theta}$. Compared to traditional supervised learning methods, prototype learning requires less labeled data and has stronger generalization ability.

However, simply aggregating all learned class prototypes using traditional prototype-based methods may render some prototypes indistinguishable from one another. To address this problem,~\citet{zhengFewShotClassIncrementalLearning2021} introduced the class structure regularizer to regulate the distribution of the learned classes in the embedding space of FSCIL. By using class distribution as prior knowledge to regularize the learning of new classes, this approach ensures that classes from the same or different sessions are distinguishable from one another.

In FSL, prototype-based methods face challenges in IL scenarios, primarily due to two issues: (i) With the increase in data volume, sample features or label distributions change because of potential concept drift or data distribution drift, making prototype samples fail to accurately represent the latest data distribution; (ii) Newly introduced later-task classes might differ conceptually from earlier classes, causing conflicts within the prototype space, thereby affecting the efficacy of prototype distance measurement and consequently influencing classification accuracy. To address these issues,~\citet{zhu2021self} proposes an incremental prototype learning scheme consisting of random episode selection and dynamic relation projection. Random episode selection improves the extensibility of the feature representation by adapting gradients to different simulated incremental processes generated randomly. Dynamic relation projection utilizes the relationship matrix between new class samples and old class prototypes to update existing prototypes.

Learning Vector Quantization (LVQ) is a prototype clustering method that selects vector points as prototypes based on distance as the clustering criterion.~\citet{chenINCREMENTALFEWSHOTLEARNING2021} uses a non-parametric method based on LVQ in deep embedding space. They compress the information of the learning task into a few quantized reference vectors. These include within-class variation, less forgetting regularization, and calibrated reference vectors to alleviate catastrophic forgetting.
Based on the idea of the CIL algorithm,~\citet{mazumderFewShotLifelongLearning2021} proposes few-shot lifelong learning (FSLL). This algorithm selects some parameters to update in each incremental session to resist overfitting. At the same time, it minimizes the cosine similarity between the new class prototypes and old class prototypes to maximize their separation, thereby improving classification performance.

According to~\citet{herscheConstrainedFewShotClassIncremental2022}, the input images are mapped to quasi-orthogonal prototypes from the perspective of hyperdimensional computing. The proposed C-FSIL comprises a frozen meta-learned feature extractor, a trainable fixed-size fully connected layer, and a rewritable dynamically growing memory. The three parameter update forms provided effectively balance accuracy and compute-memory cost.
In~\citet{yao2022few}, a human cognition-inspired prototype representation enhancement scheme is proposed for FSCIL. This method uses prototype representations and iteratively learns the knowledge of novel classes by exploring similarity correlations with previously learned classes.
~\citet{yang2023neural} argue that misalignment between the feature and classifier of old classes caused by fine-tuning the backbone or previous classifier prototypes is the reason for forgetting. Inspired by the neural collapse theory, they align a set of prototypes during neural collapse with prototypes required for FSL, which improves the classifier's performance.

The aforementioned methods exhibit conciseness in their algorithms, but the semantic gap between the few-shot class prototypes and the real data distribution is a major obstacle to improving the accuracy of prototype-based methods.

\subsubsection{Meta process}
Inspired by the multi-task optimization method MAXL~\citep{liu2019self},~\citet{chiMetaFSCILMetaLearningApproach2022} proposed MetaFSCIL, which directly transforms adapting to new knowledge and retaining old knowledge into a meta-objective. They mimicked the scenario during meta-testing by sampling a sequence of incremental tasks from base classes. Furthermore, they proposed a bi-directional guided modulation based on meta-learning to automatically adapt to new knowledge.
Drawing on metric learning within the context of meta-learning,~\citet{zou2022marginbased} discovered that using large margin classification improves the performance of the base classes but leads to a decrease in performance when learning new classes, a phenomenon termed class-level overfitting. The authors explain that this is due to the easily satisfied constraint of learning shared or class-specific patterns. Subsequently, they propose the boundary-based CLOM framework, which introduces an additional constraint that effectively addresses the aforementioned issue.

\subsection{Feature and feature space-based methods}
\subsubsection{Feature decoupling}
Feature decoupling, which entails dividing features into distinct representations, allows models to concentrate on more pertinent information. According to~\citet{zhaoMgSvFMultiGrainedSlow2021}, the disentanglement of features results in low-frequency components playing a more significant role in preserving old knowledge. Specifically, they employed discrete cosine transform to disentangle features and proposed a frequency-aware regularization method to enhance inter-space learning performance. Moreover, the proposed feature space composition operation further improves the inter-space learning performance.

\subsubsection{Feature space}
The representation of subspaces increases the efficiency of algorithms by mapping the original data to a low-dimensional space while preserving its useful features. Based on subspace representation, FSCIL projects new-class data into the subspace composed of base or old-class features, thereby enabling the model to better adapt to new classes. In~\citet{cheraghian2021synthesized}, a mixture of subspaces is proposed to describe the visual and semantic domain distribution of the data, which helps to avoid forgetting old classes. Additionally, a variational autoencoder is utilized to generate synthesized visual samples that enhance the performance of pseudo-features and prevent overfitting during IL of new classes.
In~\citet{akyurekSubspaceRegularizersFewShot2022}, the authors propose a subspace regularization scheme that encourages the weight representation of new classes to be close to the subspace spanned by the weights of existing old classes. This regularization term is straightforward and user-friendly, and can incorporate more prior knowledge.
From the perspective of parameter feature space,~\citet{kim2023warping} proposed WaRP by fusing the advantages of F2M~\citep{shiOvercomingCatastrophicForgetting2021} for finding flat minimums of the loss function and FSLL~\citep{mazumderFewShotLifelongLearning2021} for parameter fine-tuning. 
They seek directions in the parameter space that are flat with respect to the loss function, and use the method of singular value decomposition to represent the parameter space. In each incremental session, they fine-tune unimportant parameters in the parameter space to learn novel classes.

Recently, ~\citet{song2023learning} presents the concept of fantasy space to enhance semantic knowledge. The core idea is to introduce placeholders for unseen classes within the fantasy space. These placeholders derive from the original classes using discrete transformation. By learning to recognize and contrast in the fantasy space fostered by virtual classes, it boosts base classes separation and novel classes generalization.
\subsubsection{Prospective Learning}
Backward compatibility is an issue that requires special consideration in the process of software updates. It demands that newer versions of software be able to accept data from previous versions. Conversely, forward compatibility requires that older versions of software be able to accept data from newer versions. From this perspective, the ability of the FSCIL model to overcome forgetting represents its backward compatibility~\citep{zhouForwardCompatibleFewShot2022a}. This means that a model trained on a new session should not forget old class samples. Few studies have addressed the model's forward compatibility, which involves preparing for possible novel classes and updates during current training sessions. Here, we define:

\begin{definition}
\label{def-Prospective_Learning}
\textbf{Prospective Learning} \textit{refers to a certain method or technique in FSCIL, where the model is trained on base dataset to have forward compatibility performance, thus enabling the model to better handle incremental few shot novel classes.}
\end{definition}

In order to enable the model to handle new classes,~\citet{zhouForwardCompatibleFewShot2022a} proposed forward compatible training (FACT), which allocates multiple virtual prototypes as a reserved space in the feature space to make the model scalable. FACT optimizes virtual prototypes to minimize intra-class distances and reserves more space for upcoming new classes. The model is made prospective through instance mixing to generate virtual instances. In subsequent research,~\citet{zhouFewShotClassIncrementalLearning2022} proposed LIMIT, which creates fake FSCIL tasks from the base dataset and obtains generalizable features through meta-learning from different fake tasks to prepare the model for real FSCIL tasks. Additionally, an instance-specific embedding is generated by a transformer-based meta-calibration module to further improve performance.
From the perspective of open-set recognition,~\citet{pengFewShotClassIncrementalLearning2022} linked FSCIL with open-set tasks to prepare the model for new classes. Specifically, they proposed using angular penalty loss in face recognition to obtain good clustering features instead of cross-entropy loss. They combined class enhancement and data augmentation to improve the feature extractor's generalization ability for future incremental classes.

\subsection{Replay-based methods}
Based on the rehearsal technique, FSCIL approaches replay previously learned information for the task solver when presented with a new task. 
Replay-based methods employ episodic memory $\mathcal{M}$ to replay the examples from previous tasks while updating the model with the current task $t$.
There are two types: direct replay involves saving examples from old tasks to $\mathcal{M}$, while generative replay involves using a generative model to remember the distribution of data from old tasks and generate examples to $\mathcal{M}$. 
When fine-tuning the model with data $D^t$, the loss function can be expressed as:
\begin{equation} \label{eq:replay-loss}
\mathcal{L}=\frac{1}{|D^{\left( t \right)}\cup \mathcal{M}^{\left( t \right)}|}\sum_{\left( x,y \right) \in \left( D^{\left( t \right)}\cup \mathcal{M}^{\left( t \right)} \right)}{\ell \left( f\left( x \right) ,y \right)},
\end{equation}

\subsubsection{Direct replay} 
% direct replay
~\citet{kuklevaGeneralizedIncrementalFewShot2021} proposed a three-stage framework, wherein the first two stages train the network on base and novel classes separately and employ a model parameter constraint method to prevent forgetting of old classes. In the third stage, a small set of stored samples are used for replay and calibration of the classifier's performance across all classes (both base and novel classes).
IL methods based on knowledge distillation usually store a set of old class exemplars and add additional distillation loss to transfer and preserve old knowledge~\citep{rebuffi2017icarl,castro2018end,wu2019large,hou2018lifelong}. However, due to class imbalance in few-shot scenarios and performance trade-offs between novel and base classes~\citep{hou2019learning}, knowledge distillation is not the preferred method for FSCIL.~\citet{cheraghianSemanticAwareKnowledgeDistillation2021} proposed the semantic-aware knowledge distillation method by storing a small number of samples for the previous classes. By incorporating word embeddings as auxiliary information and mapping images to vector space, the effectiveness of knowledge distillation for FSCIL has been demonstrated.
Unlike CIL based on individual knowledge distillation~\citep{park2019relational},~\citet{dongFewShotClassIncrementalLearning2021a} applied graph distillation techniques to FSCIL for the first time. They proposed a scheme for exemplar relation distillation incremental learning (ERDIL) based on graph relation knowledge distillation for knowledge extraction and representation. It effectively transfers old knowledge to the model for learning new tasks by maintaining a graph that represents the relationship between classes.

\subsubsection{Generative replay}
In light of the privacy issues caused by storing real old data,~\citet{liuFewShotClassIncrementalLearning2022} proposes a data-free replay scheme for synthesizing old samples. By imposing entropy regularization, the generator is encouraged to produce uncertain examples that are closer to the decision boundary.
Since the traditional generative replay paradigm in CIL cannot be applied to FSCIL,~\citet{agarwal2022semantics} proposes few-shot incremental learning GAN (FSIL-GAN), which consists of a pre-trained feature extractor, a generator, a discriminator, and a semantic projection module. This is used to address the problem of approximating the real data distribution with a small amount of data. They first match class-specific synthesized visual features with their respective latent semantic vectors, and then ensure the diversity and distinguishability of the synthetic features through an anti mode-collapse regularizer. However, this method's performance cannot be guaranteed for multi-domain data.

\subsection{Dynamic network structure-based methods}
Dynamic network structures~\citep{sabour2017dynamic, chen2020dynamic} enable automatic adjustment of network architecture during runtime, based on input data features, thereby possessing strong generalization capabilities and reduced risks of overfitting. Due to their flexibility and robust scalability, dynamic architectures have been extensively researched for their applications in IL~\citep{rusu2016progressive, aljundi2017expert, rosenfeld2018incremental}. Leveraging these advancements, researchers have recently applied dynamic network structures in the context of FSCIL. Depending on the initial network structure employed, these methods can be categorized into three distinct groups.
\begin{figure*}[h]%
    \centering
    \includegraphics[]{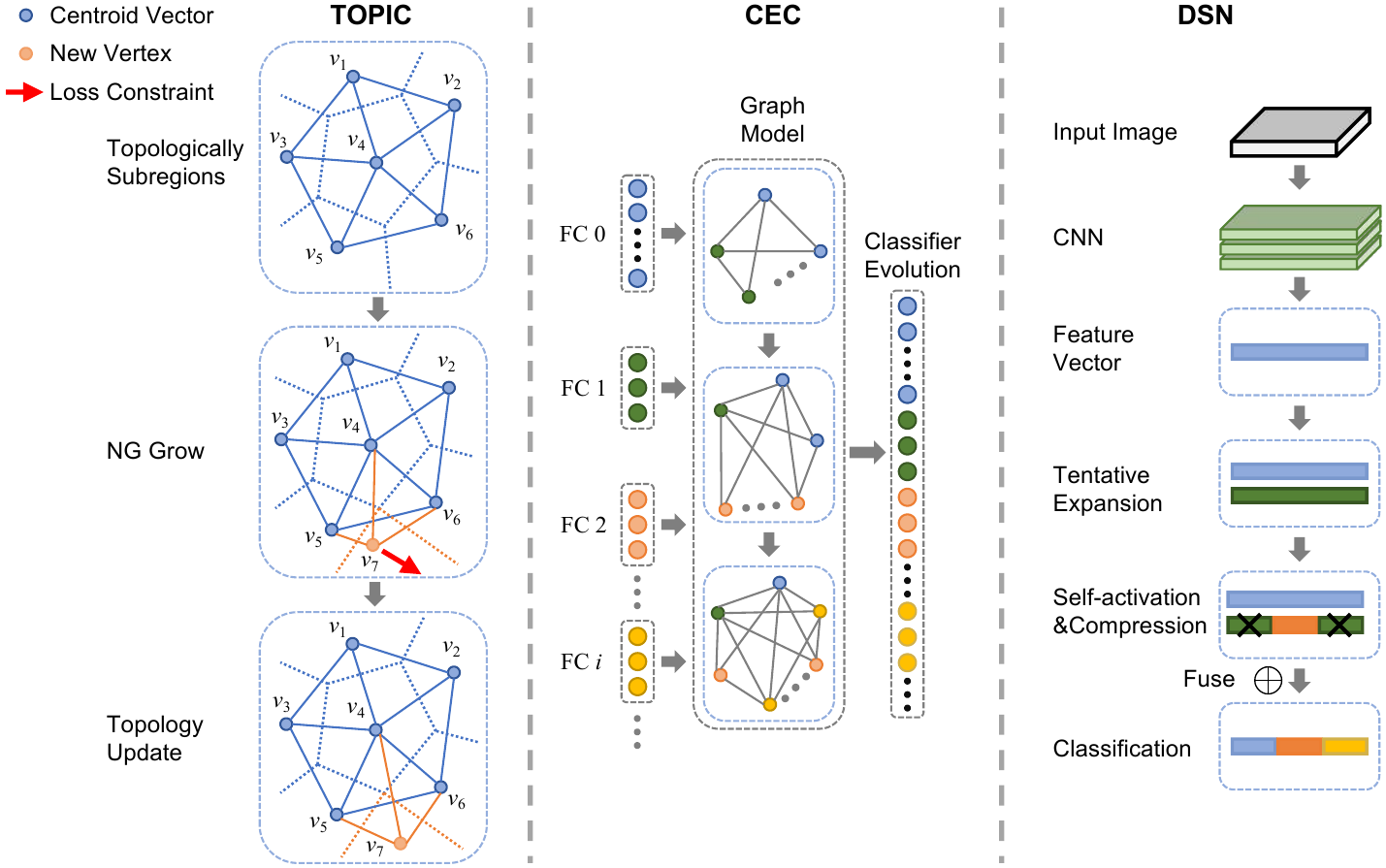}
    \caption{During training, the network structure dynamically adjusts. \textbf{Left}: Sample features form the neural graph's topology. With new nodes added, TOPIC~\citep{taoFewShotClassIncrementalLearning2020a} uses loss constraints for topology updates. \textbf{Middle}: To make the classifier suitable for all categories, CEC~\citep{zhang2021few} applies graph models to the classifier. As new tasks emerge and categories increase, the classifier's topology continuously evolves. \textbf{Right}: When training on new classes, DSN~\citep{yang2022dynamic} temporarily expands network nodes to learn new class features, and then compresses redundant nodes to provide a compact feature representation.}
    \label{fig-dynamic-network}
\end{figure*}

\subsubsection{Neural gas network}
~\citet{taoFewShotClassIncrementalLearning2020a} proposed the TOPIC framework, which utilizes a neural gas (NG) network to learn the topological structure of the feature space formed by different categories for knowledge representation. The stability of the NG's topology is maintained to prevent forgetting of old categories. With the dynamic growth of NG to accommodate new samples, the representation of few-shot new classes is improved. Fig. \ref{fig-dynamic-network} (left) displays the stabilization and adaptation of TOPIC.
\subsubsection{Graph attention network}
The Graph attention network can dynamically process different types of graph data and make dynamic decisions based on the importance of nodes and edges learned on the graph.~\citet{zhang2021few} have pointed out that decoupling the training process into embedding learning and classifier learning can effectively prevent knowledge forgetting in the backbone. They proposed the Continually Evolved Classifier (CEC), which first trains the backbone with base data to give the network strong feature extraction capabilities. Then, the graph attention model is introduced, and the graph attention network is used in the classifier layer to adapt to the changes of incremental tasks. With the arrival of incremental tasks, the nodes and weights of the Graph model dynamically increase. Fig. \ref{fig-dynamic-network} (middle) illustrates the continual evolution of classifier.
\subsubsection{Dynamic neural networks}
~\citep{yangLearnableExpansionandCompressionNetwork2021} proposed a learnable expansion-and-compression network (LEC-Net) which enhances the feature representation capability by selectively expanding the network nodes and reduces feature drift from a model regularization perspective. Furthermore, they introduce the dynamic support network (DSN)~\citep{yang2022dynamic} which can adaptively expand the network. DSN leverages compressive network expansion to enrich feature representation in each incremental task and dynamically adjusts the feature space by invoking the old class distribution. During each training, DSN selectively expands the network nodes to enhance the feature representation capability of incremental classes. Then, it dynamically compresses and expands the network through node self-activation to pursue a compact feature representation, thereby alleviating overfitting. Fig. \ref{fig-dynamic-network} (right) shows the expansion and compression of DSN.

In the latest study,~\citet{yoon2023soft} explores a masking-based method in network structure. They utilize non-binary masks to construct soft-subnetworks from the original network, effectively balancing forgetting and overfitting. In the base classes session, soft-subnetwork parameters and weight score are learned. In the incremental learning session, minor parameters of the subnetwork are updated.

\subsection{Methods summary}
This section reviews recent advancements in FSCIL. The following critically examines the strengths and weaknesses of various families.

Traditional machine learning methods offer promising research prospects. By carefully designing the supervised approach of the model, introducing additional data proves effective. Studying FSCIL from a statistical distribution or function optimization perspective enhances model interpretability. However, the complexity of statistical distribution modeling still presents difficulties.

Meta learning-based methods aim to make machine learning models more flexible and adaptive. But meta-learning typically assumes all tasks are from the same or similar data distributions and has high dependence on the meta-training set. When incremental tasks have different distributions from the base classes, model performance can be affected.

Feature and feature space-based methods leverage the core idea of learning more robust and efficient feature representations. In particular, prospective learning methods are worth exploring for their natural capability in handling unseen samples.

Replay-based methods directly address catastrophic forgetting in FSCIL. However, direct replay faces constraints in storage space, sample selection, and privacy. In contrast, generative replay partially alleviates these issues and offers a more flexible approach. Nevertheless, the challenges of training complexity and subpar data quality persist in generative replay methods.

Dynamic network structure-based methods serve as vital solutions to FSCIL challenges. They adapt to continuously changing data streams by adjusting model structures or inter-class relationships, thereby learning new knowledge while retaining old knowledge. Dynamic Networks have gained traction in IL~\citep{wang2022dualprompt,wang2022foster}, and exploring their application in FSCIL is encouraged.

Overall, there remains an open research challenge to develop methods that harmoniously balance performance, scalability, efficiency, and complexity.

\section{Model performance} \label{sec-5}
In this section, we will present the performance of typical FSCIL methods on three different datasets. Firstly, we will outline the methodology for model selection, followed by an introduction to the classical datasets and evaluation metrics. Finally, we will summarize the performance results of various models.

\subsection{Model selection}
Comparing the performance of different methods is necessary, but currently, many of these methods' codes are not publicly available. As most studies follow the standards set forth by~\citet{taoFewShotClassIncrementalLearning2020a} (see Section \ref{sec:Problem formalization}), it is feasible to use the data reported in the original papers of the methods being compared, and we have adhered to this principle. Thus, the results reported in this  section are based on the original paper's reported data or the data processed from these original data. We have selected and compared the performance of 22 methods from five different families.

\subsection{Datasets}
At present, there is no specific dataset for FSCIL, and most of them are made from existing datasets for new tasks.
In the majority of FSCIL experiments~\citep{taoFewShotClassIncrementalLearning2020a,zhu2021self,cheraghianSynthesizedFeatureBased2021,zhang2021few,liuFewShotClassIncrementalLearning2022,chiMetaFSCILMetaLearningApproach2022,zhengFewShotClassIncrementalLearning2021,zhouFewShotClassIncrementalLearning2022,pengFewShotClassIncrementalLearning2022,shiOvercomingCatastrophicForgetting2021,yang2022dynamic}, the three image classification datasets  CIFAR-100~\citep{krizhevsky2009learning}, MiniImageNet~\citep{vinyals2016matching} and CUB-200~\citep{wah2011caltech} are commonly used. 

\textbf{CIFAR-100} contains 100 classes with 600 RGB images per class, where each class has 500 training images and 100 testing images. The size of each image is $32\times32$ pixels.

\textbf{MiniImageNet} contains 60000 RGB images of size $84\times84$ pixels from ImageNet-1k~\citep{deng2009imagenet}. It possesses the same number of classes and samples as CIFAR-100, but its content is more complex and valuable for FSCIL research.

\textbf{CUB-200} is currently the most widely used benchmark image dataset for fine-grained classification and recognition research. The dataset has a total of 11,788 bird images, including 200 bird subclasses, of which the training dataset has 5,994 images and the test set has 5,794 images. Each image has a size of 224 × 224 pixels. It provides more sessions and incremental classes for comparing the sensitivity of different methods.

The performance of the selected method was evaluated on the three benchmark datasets mentioned above. For detailed dataset settings refer to Table \ref{tab:three-datasets}.

\begin{table}[]
\centering
\renewcommand\arraystretch{1.5} % 增加行高
\fontsize{7pt}{6pt}\selectfont
\caption{Experimental setup for the three datasets}
\label{tab:three-datasets}
\begin{tabular}{cccc}
\toprule
\textbf{Dataset}                                & \textbf{Base classes} & \textbf{Incremental sessions setup} & \textbf{Sessions} \\ \midrule
CIFAR-100  & 60           & 5-way, 5-shot                                                        & 8        \\
MiniImageNet & 60           & 5-way, 5-shot                                                        & 8        \\
CUB-200            & 100          & 10-way, 5-shot                                                       & 10       \\ \bottomrule
\end{tabular}
\end{table}

% \begin{table}[]
% \centering
% \renewcommand\arraystretch{1.5} % 增加行高
% \fontsize{7pt}{6pt}\selectfont
% \caption{Experimental setup for the three datasets}
% \label{tab:three-datasets}
% % \begin{tabular}{m{2.5cm}<{\centering}m{1.3cm}<{\centering}m{2.1cm}<{\centering}m{0.8cm}<{\centering}}
% \begin{tabular}{m{2.9cm}<{\centering}m{1.0cm}<{\centering}m{2.1cm}<{\centering}m{0.7cm}<{\centering}}
% \toprule
% \textbf{Dataset}                                & \textbf{Base classes} & \textbf{Incremental sessions setup} & \textbf{Sessions} \\ \midrule
% CIFAR-100~\citep{krizhevsky2009learning}  & 60           & 5-way, 5-shot                                                        & 8        \\
% MiniImageNet~\citep{vinyals2016matching} & 60           & 5-way, 5-shot                                                        & 8        \\
% CUB-200~\citep{wah2011caltech}            & 100          & 10-way, 5-shot                                                       & 10       \\ \bottomrule
% \end{tabular}
% \end{table}

\subsection{Metrics}
Considering the scarcity of original data reported in the paper, we solely compared the accuracy of each session, average accuracy (AA) of all sessions and Performance dropping rate (PD)~\citep{zhang2021few}. PD measures the absolute accuracy drops in the last session w.r.t. the accuracy in the base session, defined as
\begin{equation} \label{eq:PD}
\text{PD}=\mathcal{A}_0-\mathcal{A}_N,
\end{equation}
where $\mathcal{A}_0$ is the classification accuracy in the base session and $\mathcal{A}_N$ is the accuracy in the last session.

\subsection{Results}
\subsubsection{Benchmark results}
\begin{table*}[h]
\caption{AA ($\%$) and PD ($\%$) in CIFAR-100, MiniImageNet and CUB-200.}
\label{result_on_three_datasets}
\centering
\begin{threeparttable}
\renewcommand\arraystretch{2} 
\fontsize{6pt}{6pt}\selectfont
\begin{tabular}{m{2cm}<{\centering}m{3.5cm}<{\centering}m{1.6cm}<{\centering}m{0.5cm}<{\centering}m{0.5cm}<{\centering}m{0.5cm}<{\centering}m{0.5cm}<{\centering}m{0.5cm}<{\centering}m{0.5cm}<{\centering}m{0.5cm}<{\centering}m{0.5cm}<{\centering}m{0.5cm}<{\centering}}
\toprule
\multirow{2}{2cm}{\centering \textbf{Families}}                 & \multirow{2}{*}{\textbf{Methods}}            & \multirow{2}{*}{\textbf{Venue}} & \multicolumn{3}{c}{\textbf{CIFAR-100}}              & \multicolumn{3}{c}{\textbf{MiniImageNet}}             & \multicolumn{3}{c}{\textbf{CUB-200}}                \\ \cline{4-12} 
                          &                    &         & \begin{tabular}[c]{@{}c@{}}  AA$\uparrow$\end{tabular} & PD$\downarrow$ & ResNet & \begin{tabular}[c]{@{}c@{}}  AA$\uparrow$\end{tabular} & PD$\downarrow$ & ResNet & \begin{tabular}[c]{@{}c@{}}  AA$\uparrow$\end{tabular} & PD$\downarrow$ & ResNet \\ \midrule
\multirow{5}{2cm}{\centering Traditional machine learning methods}  & SSFSCIL~\citep{cui2021semi}          & ICIP 2021      & -                & -    & -  & -                & -    & -  & 50.78               & 34.64   & 18  \\
                          & GP-Tree~\citep{achituveGPTreeGaussianProcess2021}    & ICML 2021      & -                & -    & -  & -                & -    & -  & 54.26               & 30.12   & 18  \\
                          & F2M~\citep{shiOvercomingCatastrophicForgetting2021}    & NeurIPS 2021      & 53.65               & 20.04   & 18  & 54.89               & 22.63   & 18  & 69.49               & 20.81   & 18  \\
                          & LDC~\citep{liuLearnableDistributionCalibration2022}    & arXiv 2022      & -                & -    & -  & -                & -    & -  & 68.32               & 16.31   & 18  \\
                          & FeSSSS~\citep{ahmadFewShotClassIncremental2022}     & CVPR 2022      & -                & -    & -  & 68.24               & 22.63   & 18  & 62.86               & 26.62   & 18  \\ \midrule
\multirow{7}{2cm}{\centering Meta learning-based methods}    & FSLL~\citep{mazumderFewShotLifelongLearning2021}    & AAAI 2021      & -                & -    & -  & -                & -    & -  & 62.62               & 19.81   & 18  \\
                          & SPPR~\citep{zhuSelfPromotedPrototypeRefinement2021}    & CVPR 2021      & 54.51               & 20.85   & 18  & 52.75               & 19.53   & 18  & 49.32               & 31.35   & 18  \\
                          & CSR~\citep{zhengFewShotClassIncrementalLearning2021}   & ICDMW 2021      & 59.07               & 23.02   & 20  & 54.11               & 23.15   & 18  & 62.32               & 19.60   & 18  \\
                          & C-FSCIL~\citep{herscheConstrainedFewShotClassIncremental2022} & CVPR 2022      & 61.64               & 27.00   & 12  & 61.61               & 24.99   & 12  & -                & -    & -  \\
                          & MetaFSCIL~\citep{chiMetaFSCILMetaLearningApproach2022}   & CVPR 2022      & 60.79               & 24.53   & 20  & 58.85               & 22.85   & 18  & 61.93               & 23.26   & 18  \\
& CLOM~\citep{zou2022marginbased} & NeurIPS 2022 & 60.57	& 23.95 & 20 & 58.48	& 25.08	& 18 & 67.17	& 19.99	& 18\\ 					 & NC-FSCIL~\citep{yang2023neural}         & ICLR 2023      & 67.50                & 26.41   & 12  & 67.82               & 25.71   & 12  & 67.28               & 21.01   & 18  \\ \midrule
\multirow{5}{2cm}{\centering Feature space and feature-based methods} & VAE-based*~\citep{cheraghian2021synthesized}      & ICCV 2021      & 50.86               & 20.36   & 18  & 50.63               & 19.30   & 18  & 51.84               & 25.55   & 18  \\
                          & FACT~\citep{zhouForwardCompatibleFewShot2022a}     & CVPR 2022      & -                & -    & -  & -                & -    & -  & 64.42               & 18.96   & 18  \\
                          & ALICE~\citep{pengFewShotClassIncrementalLearning2022}   & ECCV 2022      & 63.21               & 24.90   & 18  & 63.99               & 24.90   & 18  & 65.75               & 17.30   & 18  \\
                          & LIMIT~\citep{zhouFewShotClassIncrementalLearning2022}   & PAMI 2022      & 61.84               & 22.58   & 20  & 59.06               & 23.13   & 18  & 65.48               & 18.48   & 18  \\
                          & MgSvF~\citep{zhaoMgSvFMultiGrainedSlow2021}      & PAMI 2022      & -                & -    & -  & -                & -    & -  & 62.37               & 17.96   & 18  \\ \midrule
Replay-based methods          & ERDR~\citep{liuFewShotClassIncrementalLearning2022}    & ECCV 2022      & 60.77               & 24.26   & 20  & 58.02               & 23.63   & 18  & 61.52               & 23.51   & 18  \\ \midrule
\multirow{4}{2cm}{\centering Dynamic network structure-based methods} & TOPIC~\citep{taoFewShotClassIncrementalLearning2020a}   & CVPR 2020      & 42.62               & 34.73   & 18  & 39.64               & 36.89   & 18  & 43.92               & 42.40   & 18  \\
                          & CEC~\citep{zhang2021few}          & CVPR 2021      & 59.53               & 23.93   & 20  & 57.75               & 24.37   & 18  & 61.33               & 23.57   & 18  \\
                          & LEC-Net~\citep{yangLearnableExpansionandCompressionNetwork2021} & arXiv 2022      & 43.14               & 29.37   & 18  & -                & -    & -  & 45.09               & 38.90   & 18  \\
                          & DSN~\citep{yang2022dynamic}      & PAMI 2022      & 60.14               & 23.00   & 18  & 54.39               & 21.06   & 18  & 71.02               & 17.65   & 18  \\ \bottomrule
\end{tabular}
\begin{tablenotes}
      \item * The method name "VAE-based" is defined by us.
\end{tablenotes}
\end{threeparttable}
\end{table*}

\begin{figure*}[h]%
    \centering
    \includegraphics[scale=0.9]{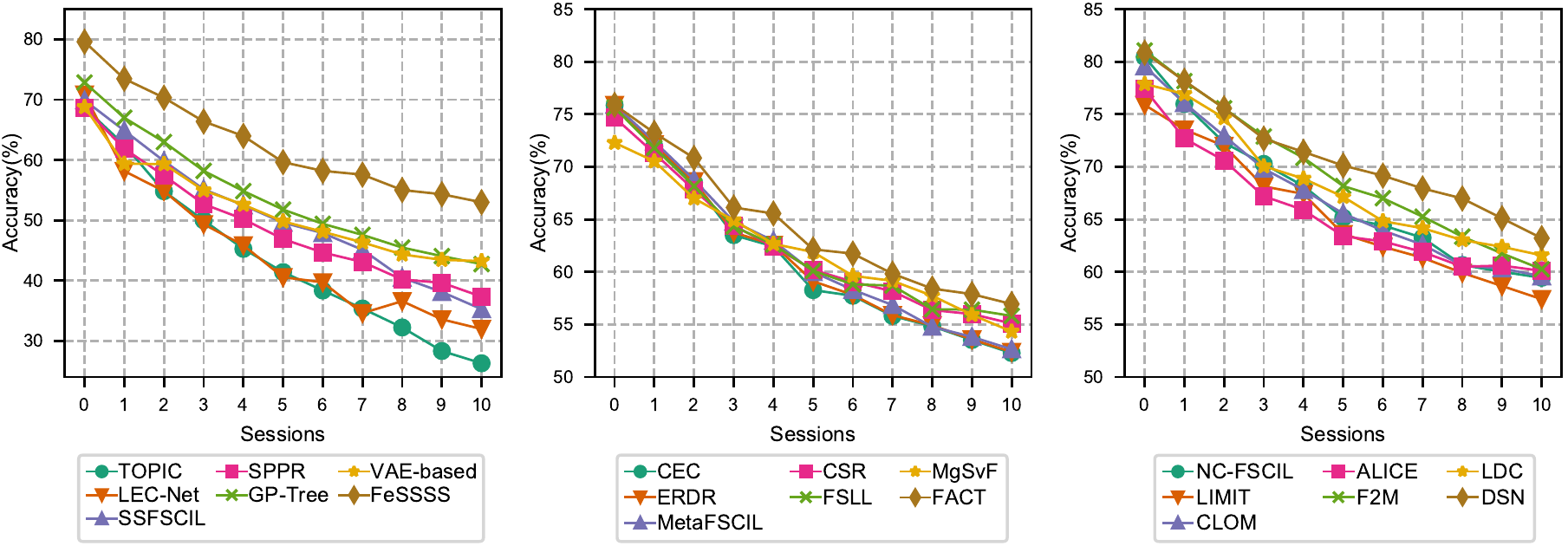}
    \caption{Accuracy curves of different methods on each session of CUB-200 dataset.}
    \label{curves on cub}
\end{figure*}

\textbf{Average performance.} Table \ref{result_on_three_datasets} presents the performance of typical FSCIL methods on different datasets. 
In the comparative experiments, all methods utilized ResNet as the backbone. However, there were variations in the specific ResNet models used (e.g., ResNet-12, ResNet-18, ResNet-20). These differences are detailed in the table.
We observe substantial performance disparities among different methods for various datasets. For the small-sized CIFAR-100 dataset, NC-FSCIL~\citep{yang2023neural} exhibits outstanding performance at $67.50\%$, outperforming other methods by a large margin. For the more challenging MiniImageNet dataset, FeSSSS~\citep{ahmadFewShotClassIncremental2022} utilizes self-supervised learning for data augmentation and achieves a performance of $68.24\%$, surpassing NC-FSCIL~\citep{yang2023neural} while also exhibiting lower knowledge forgetting. For the fine-grained CUB-200 dataset, only DSN~\citep{yang2022dynamic} with AA surpasses $70\%$ with a performance of $71.02\%$, demonstrating a better ability to capture the differences between categories.

\textbf{Performance comparison by session.} 
The accuracy of each session during the incremental process of various models on the CUB-200 dataset is illustrated in the line chart in Fig. \ref{curves on cub}. The accuracy of the model on the base classes limits the accuracy improvement during the incremental phase. With the exception of some early methods (TOPIC~\citep{taoFewShotClassIncrementalLearning2020a}, SPPR~\citep{zhuSelfPromotedPrototypeRefinement2021}, VAE-based~\citep{cheraghian2021synthesized}), most methods have an accuracy of $70\%$ to $80\%$ on the base dataset, and few methods have an accuracy above $80\%$ on the base dataset (F2M~\citep{shiOvercomingCatastrophicForgetting2021}, DSN~\citep{yang2022dynamic}, NC-FSCIL~\citep{yang2023neural}). As the earliest research, TOPIC~\citep{taoFewShotClassIncrementalLearning2020a} was no longer competitive in each session of the training. F2M~\citep{shiOvercomingCatastrophicForgetting2021} based on function optimization and DSN~\citep{yang2022dynamic} based on dynamic neural networks still demonstrate high performance advantages.
 
\subsubsection{Performance comparison of accuracy and inference speed}
\begin{figure}[h]% 
    \centering
    \includegraphics[]{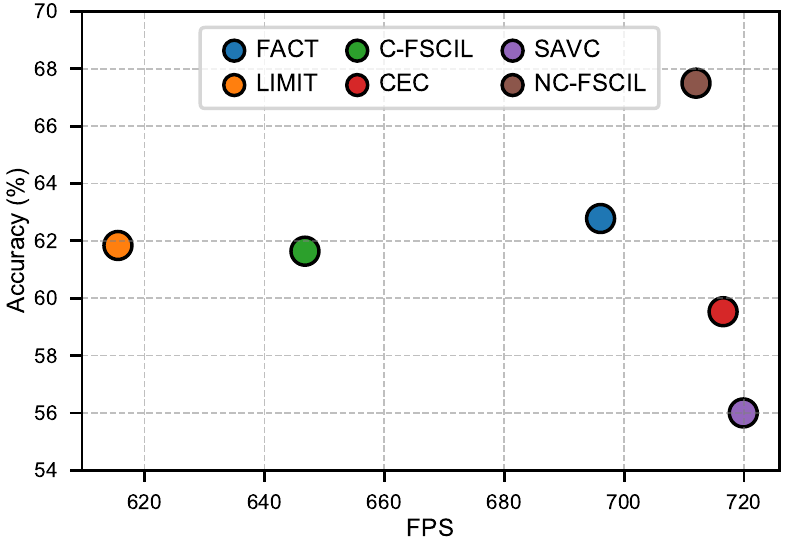}
    \caption{Performance comparison of various methods on CIFAR100: FPS vs. Accuracy}\label{fig-scatter}
\end{figure}

Due to the unavailability of source code for most methods, in this part, we only select methods with publicly available code. We test the accuracy and inference speed of these methods on the CIFAR100 dataset. All experiments are conducted 50 times on an NVIDIA TITAN V GPU with 12GB of memory, and the average values are reported as the final results. The experimental results are presented in Fig. \ref{fig-scatter}. It is noticeable that NC-FSCIL~\citep{yang2023neural}, based on the neural collapse theory, leads in both accuracy and inference speed. While SAVC~\citep{song2023learning} and CEC~\citep{zhang2021few} methods exhibit lower accuracy, they benefit from reduced model complexity, achieving the fastest inference speeds.

\section{Research on few-shot incremental learning applications} \label{sec-6}
In Section \ref{sec-3}, the focus lies on fundamental research in FSCIL. In this section, we primarily introduce research that concentrates on implementing FSCIL techniques to resolve practical predicaments. We do not distinguish between the variants of FSCIL, such as FSCIL and FSIL, but rather focus on their applications. FSCIL, originating from computer vision (CV), has presently gained extensive usage in natural language processing (NLP) and Graph technology as well. Further subdivisions can be observed in Table \ref{tab:yingyong}.

\begin{sidewaystable*}[htp]
\centering
\begin{threeparttable}
\caption{Summary of FSCIL applied research, including CV, NLP, and graph domains.}
\label{tab:yingyong}
\renewcommand\arraystretch{1.5}
\fontsize{6pt}{6pt}\selectfont

\begin{tabular}{m{1.1cm}<{\centering}|m{3.cm}<{\centering}|m{3.8cm}<{\centering}m{6.5cm}<{\raggedright}m{4cm}<{\centering}m{2.5cm}<{\centering}}
\toprule
\multicolumn{1}{c|}{\textbf{Domain}} & \textbf{Scenario}                                              & \multicolumn{1}{c}{\textbf{Method}}                                     & \multicolumn{1}{c}{\textbf{Content}}                                                          & \multicolumn{1}{c}{\textbf{Dataset}}                                                                          & \multicolumn{1}{c}{\textbf{Metrics}} \\ \hline
\multirow{27.5}{*}{\textbf{CV}}        & \multirow{1}{3cm}{\centering Hyperspectral image} & LPILC~\citep{bai2020class}                                                & Propose a linear programming IL classifier                                  & PaviaU, Indian Pines,...                                                                                      & AA                          \\ \cline{2-6} 
                                     &       \multirow{1}{3.1cm}{\centering Pedestrian attribute recognition}   & APGM~\citep{xiang2019incremental}                                         & Incremental few-shot learning for pedestrian attribute recognition                            & PETA~\citep{deng2014pedestrian}, RAP~\citep{li2016richly}                                                         & mA,Precision,Recall,F1      \\ \cline{2-6} 
                                     & \multirow{4.5}{2.8cm}{\centering Fine-grained image}  & MetaSearch~\citep{wang2020metasearch}                                     & Few-shot incremental fine-grained product search                                              & Mini-ImageNet~\citep{vinyals2016matching},RPC~\citep{wei2019rpc}                                                  & Accuracy                    \\
                                     &                                                                & CSFL~\citep{li2022few}                                                    & FSCIL for fine-grained vehicle recognition                                                    & Stanford Cars~\citep{Krause_2013_ICCV_Workshops}, CompCars~\citep{Yang_2015_CVPR}                                 & Accuracy,AA,PD              \\
                                     &                                                                & SSFE-Net~\citep{pan2023ssfe}                                              & A self-supervised approach is proposed for ultra-fine-grained FSCIL                           & Cotton~\citep{yu2021benchmark},SoyCultivarLocal~\citep{yu2021benchmark},...                               & Accuracy                    \\ \cline{2-6} 
                                     & \multirow{8}{2.8cm}{\centering Object detection}                   & ONCE~\citep{perez2020incremental}                                         & The first study on incremental few-shot object detection                                      & Pascal-VOC,COCO                                                                                               & AP, AR                      \\
                                     &                                                                & Meta-Learning-Based*~\citep{chengMetaLearningBasedIncrementalFewShot2022} & Introducing meta-learning and redesigning the base framework                                  & Pascal-VOC,COCO                                                                                               & AP, AR                      \\
                                     &                                                                & Sylph~\citep{yinSylphHypernetworkFramework2022}                           & Designing a class-conditional hypernetwork for incremental few-shot object detection          & COCO,LVIS                                                                                                     & AP                          \\
                                     &                                                                & Incremental-DETR~\citep{dongIncrementalDETRIncrementalFewShot2022}        & DETR method based on fine-tuning and self-supervised learning                                 & Pascal-VOC,COCO                                                                                               & AP                          \\
                                     &                                                                & MCH, BPMCH~\citep{feng2022incremental}                                    & Analogous to the maintenance of new knowledge by establishing new connections in human cells  & Pascal-VOC,COCO                                                                                               & AP, AR                      \\ \cline{2-6} 
                                     & \multicolumn{1}{c|}{Road object detection}                     & DualFusion~\citep{tambwekarFewShotBatchIncremental2021}                   & Few-shot batch incremental road object detection                                              & IDD~\citep{varma2019idd},COCO                                                                                   & AP                          \\ \cline{2-6} 
                                     & \multicolumn{1}{c|}{Surface defect detection}                  & DKAN~\citep{sun2022new}                                                   & Knowledge distillation network for incremental few-shot surface defect detection              & NEU-DET~\citep{song2013noise}                                                                                   & AP                          \\ \cline{2-6} 
                                     & \multirow{2.8}{2.8cm}{\centering Semantic segmentation}              & PIFS~\citep{cermelli2021prototype}                                        & Prototype-based incremental few-shot semantic segmentation                                    & Pascal-VOC 2012~\citep{pascal-voc-2012}, COCO~\citep{lin2014microsoft}                                            & mIoU                        \\
                                     &                                                                & EHNet~\citep{shi2022incremental}                                          & Adaptive-update and hyper-class representation for incremental few-shot semantic segmentation & PASCAL-5~\citep{Shaban2017One},COCO                                                                             & mIoU                        \\ \cline{2-6} 
                                     & \multirow{2.8}{2.8cm}{\centering Instance segmenter}                 & iMTFA~\citep{ganea2021incremental}                                        & The first study on incremental few-shot instance segmentation                                 & Pascal-VOC~\citep{everingham2010pascal},COCO                                                                    & AP                          \\  
                                     &                                                                & iFS-RCNN~\citep{nguyen2022ifs}                                            & Incremental few-shot instance segmentation with Bayesian learning                             & COCO,LVIS~\citep{gupta2019lvis}                                                                                 & AP                          \\ \hline
\multirow{8.5}{*}{\textbf{NLP}}        & \multicolumn{1}{c|}{Intent recognition}                        & GAL~\citep{zhangIncorporatingGeometryKnowledge2022}                       & IL structure for few-shot intent recognition                                & CLINC-150~\citep{cavalin2020improving},ATIS~\citep{hemphill1990atis}                                              & Accuracy                    \\ \cline{2-6} 
                                     & \multicolumn{1}{c|}{Relation learning}                         & ERDA~\citep{qin2022continual}                                             & Continual few-shot relation learning                                                          & FewRel~\citep{han2018b},TACRED~\citep{zhang2017position}                                                          & Accuracy                    \\ \cline{2-6} 
                                     & \multicolumn{1}{c|}{Named entity recognition}                  & NER*~\citep{wangFewShotClassIncrementalLearning2022}                      & FSCIL for named entity recognition                              & CoNLL03~\citep{sang2003introduction},Ontonote 5.0~\citep{weischedel2013ontonotes}                                 & F1                          \\ \cline{2-6} 
                                     & \multicolumn{1}{c|}{Language learning}                         & LFPT5~\citep{qin2021lfpt5}                                                & Lifelong few-shot language learning based on prompt tuning                                    & CoNLL03,AGNews~\citep{zhang2015character}, CNNDM~\citep{nallapati2016abstractive},...                             & F1,Accuracy,ROUGE scores    \\ \hline
\multicolumn{1}{c|}{\textbf{NLP+CV}} & \multicolumn{1}{c|}{Label-to-image translation}                & FILIT~\citep{chenFewShotIncrementalLearning2022}                          & FSIL for label-to-image translation                                  & ADE20K~\citep{zhou2017scene},COCO-Stuff~\citep{caesar2018coco}                                                    & mIoU,accu,...               \\ \hline
\multirow{4.5}{*}{\textbf{Graph}}      & \multirow{4.5}{2.8cm}{\centering FSCIL in graph}                     & HAG-Meta~\citep{tan2022graph}                                             & The first GFSCIL research and solved based on prospective learning                            & Amazon-Clothing~\citep{mcauley2015inferring}, DBLP~\citep{tang2008arnetminer}, Reddit~\citep{hamilton2017inductive} & Accuracy,PD,RPD             \\
                                     &                                                                & Geometer~\citep{lu2022geometer}                                           & GFSCIL based on class prototype representation                                                & Cora-ML~\citep{bojchevski2018deep},Cora-Full~\citep{bojchevski2018deep},Flickr~\citep{Zeng2020GraphSAINT},...       & Accuracy                    \\ \bottomrule
\end{tabular}

\begin{tablenotes}
      \item [1] * The method name "Meta-Learning-Based" is defined by us.
      \item [2] Here are the explanations for some of the metrics:
      
        AA: Average accuracy.
        
        PD: Performance dropping rate, see Eq. \ref{eq:PD}.
        
        RPD: Relative performance dropping rate, which is the PD normalized by the initial accuracy, $RPD=PD/A_0$.
        
        mIoU: Mean Intersection-over-Union.
        
        accu: Pixel accuracy.
    \end{tablenotes}
\end{threeparttable}
\end{sidewaystable*}

\subsection{Few-shot incremental learning in computer vision}
\subsubsection{Applications in image classification}
To address the increasing demand for classification in hyperspectral imaging,~\citet{bai2020class} proposes a linear programming IL classifier.
In pedestrian attribute recognition for video surveillance, as the need for identifying new attributes increases, old models become inadequate. Based on the idea of meta-learning,~\citet{xiang2019incremental} uses an attribute prototype generator module and attribute relationship module to generate novel classification weights from annotated data.

The FSCIL method mentioned in Section \ref{sec-3} is mainly used for general classification tasks and neglects the discrimination power of learned representations, making it unsuitable for fine-grained image tasks. Based on the idea of meta-learning,~\citet{wang2020metasearch} proposes the MetaSearch model to attempt to solve the few-shot incremental product search problem in shopping and checkout processes. MetaSearch extracts different features between various novel categories to perform incremental product search. The designed multipooling-based feature extractor can capture subtle differences between fine-grained product categories, thereby improving classification accuracy.
To address the fine-grained vehicle recognition problem, a compact and separable feature learning method (CSFL) is proposed in~\citet{li2022few}. CSFL first decouples the feature extractor from the classifier and uses metric learning to train the feature extractor. In the class incremental stage, only the classifier is updated, and incremental LDA is introduced to learn intra-class compact and inter-class separable features, thereby giving the model fine-grained image recognition capabilities.
For the even more challenging ultra-fine-grained visual categorization task,~\citet{pan2023ssfe} proposes the use of self-supervised learning and knowledge distillation to enhance the feature extraction ability of the network backbone, achieving better performance on fine-grained datasets than the classic FSCIL method.

\subsubsection{Applications in object detection}
Equipping computer systems with the ability to learn from few examples for object detection has strong practical significance. Inspired by meta learning, Kang~\citep{kang2019few} proposed a novel few-shot detection model. Since the model lacks the ability to incrementally learn new targets from data streams over time, it cannot be extended to real-world deployments in open environments and edge devices.
There are also some researchers~\citep{shmelkov2017incremental,chen2019new,liu2020incdet} who study the problem of incremental target detection from the perspective of IL. But none of these methods can cope with the situation where the novel target data is few.
~\citet{perez2020incremental} introduced the incremental few-shot object detection (iFSD) paradigm, where new classes are made available gradually through different sessions. Perez proposed the ONCE model to solve the iFSD problem, which is based on the CenterNet~\citep{zhou2019objects} one-stage detection method. First, the model uses abundant base dataset to train a class-generic feature extractor. Then, a meta-learning algorithm is used to train a class-specific code generator for each novel category to register new classes. Incrementally appearing new class samples only need to be registered in the meta-training phase through forward propagation without revisiting base classes or iteratively updating, making it suitable for deployment on embedded devices.
Most subsequent methods employ class-agnostic feature extractors trained on abundant base data, following the BPNF strategy (see definition \ref{def-BPNF}), and continuously register new embeddings when novel classes emerge.~\citet{chengMetaLearningBasedIncrementalFewShot2022} also utilize CenterNet as the fundamental framework, similar to~\citet{perez2020incremental}, but introduce a novel meta-learning method for fine-tuning the model, thus retaining the knowledge related to base classes. During meta-learning optimization, they draw inspiration from the model-agnostic meta-learning algorithm~\citep{finn2017model}, a few-shot meta-learning algorithm that uses gradient descent to identify an appropriate initialization that can quickly adapt to the few samples of unseen classes. However, due to overfitting of the feature extractor on base class samples, the model's generalization of output features is inadequate, limiting the proposed model's performance on new classes.
~\citet{yinSylphHypernetworkFramework2022} proposed a hypernetwork framework for iFSD called Sylph. It uses a base detector and hypernetwork architecture similar to ONCE. Unlike ONCE, they trained a base detector with class-agnostic localization capability on abundant base dataset, thus decoupling localization from classification. This simplifies the task, but when the size of the base dataset is small or the dataset quality is poor, the class-agnostic detector's localization ability is poor. They further improved the detection accuracy by modifying the network structure and adding normalization to the predicted parameters.
~\citet{dongIncrementalDETRIncrementalFewShot2022} first introduced the DETR object detector~\citep{zhu2020deformable} into few-shot object detection and proposed Incremental-DETR. They still followed the BPNF guideline. The entire model is divided into two stages. First, the entire network is pre-trained using a large amount of data from base classes, and a self-supervised algorithm is used to fine-tune the class-specific projection layer and classification head. Then, the backbone is frozen, and the class-specific projection layer and classification head are fine-tuned for novel classes.
In contrast,~\citet{feng2022incremental} proposed a multi-class head model that mimics the mechanism of maintaining new knowledge by building new connections in human cells. The classification header is continuously added as new data appears. The classification head performs classification detection by using features learned from the data, simulating the way humans learn and maintain new knowledge. Furthermore, by adding a new backbone to the multi-class head model, a bi-path multi-class head model is formed to achieve the transfer from old knowledge to new knowledge.

In practical applications,~\citet{tambwekarFewShotBatchIncremental2021} proposed a few-shot batch incremental road object detection method specifically designed for road objects. The DualFusion architecture they proposed consists of a Faster R-CNN used for base classes detection, a novel class detection network, and a fusion network. When detecting each new class, only 10 annotated instances are used. The limitation of this method is that although access to the base dataset is only required once, all novel few-shot data must be retained to permanently access novel class data.
In the field of hot-rolled steel strip surface defects,~\citet{sun2022new} proposes a new knowledge distillation network called dual knowledge align network. Following the BPNF guidelines, a knowledge distillation framework is designed for fine-tuning. They convert NEU-DET~\citep{song2013noise} into an incremental few-shot dataset, and the experiment shows that they achieve great performance compared to other methods.
Furthermore, the few-shot incremental object learning problem for robotic vision is highly valuable. Previous studies have explored the use of a small set of visual examples to incrementally train robots and enhance their recognition capabilities~\citep{ayub2020tell}. However, the few-shot incremental object learning problem for robotic vision remains unresolved~\citep{Ayub2021FSIOL}.

\subsubsection{Applications in image segmentation}
Unlike image classification and object detection, image segmentation requires classification of each pixel, making it more challenging than the other two tasks. Instance segmentation, a subtask of image segmentation, is even more difficult than semantic segmentation as it requires distinguishing boundaries between different instances, while semantic segmentation only requires distinguishing objects and background. In the following, we will discuss some applications of FSCIL in semantic and instance segmentation.
~\citet{cermelli2021prototype} proposed the first attempt to solve incremental few-shot semantic segmentation. They proposed PIFS, which combines prototype learning with knowledge distillation. In the base stage, PIFS trains the network on base data to develop the capability of feature extraction. In the FSL stage, PIFS exploits prototypes to initialize classifiers of new classes and fine-tunes the network to refine its feature representation. The subsequently added prototype-based distillation loss enables the model to avoid overfitting and forgetting.~\citet{shi2022incremental} proposed the Embedding adaptive-update and Hyper-class representation Network (EHNet) for incremental few-shot learning. The category embedding describes exclusive semantic properties, and the hyper-class knowledge expresses class-shared semantic properties. The category embedding is stored in the memory pool and can be updated adaptively. Subsequently, in the segmentation stage, EHNet guides the query image to segment the corresponding category.

For more challenging incremental few-shot instance segmentation,~\citet{ganea2021incremental} introduced Model agnostic methods and proposed the first approach to solving this problem: iMTFA. It repurposes the Mask R-CNN network~\citep{he2017mask} to train feature extractors to generate discriminative embeddings for different instances. The average of those class embeddings is used as the representation for each class in the cosine similarity classifier. Thanks to the ability to predict localization and segmentation in a class-agnostic manner, adding new classes simply uses the representation of each class. 
When a new class appears,~\citet{nguyen2022ifs} fine-tunes the Mask-RCNN that was pre-trained on base classes. Specifically, they use Bayesian learning to estimate the class-weight distribution to modify the classification head and compute the uncertainty of prediction to modify the bounding-box head. This results in better performance than iMTFA on the COCO dataset. 
However, they do not successfully explain why their estimation of the uncertainty of bounding-box localization surpasses a Gaussian-based uncertainty estimation~\citep{he2019bounding}.

\subsection{Few-shot incremental learning in natural language processing}
FSIL is first proposed in the computer vision field, but with its increasing influence, many studies have applied its ideas to natural language processing (NLP).
For instance, in few-shot intent recognition used for text data,~\citet{zhangIncorporatingGeometryKnowledge2022} proposes constructing an undirected fully connected geometry structure based on the spatial distribution of selected samples in the embedding space. Subsequently, they apply a multisource contrastive-based loss to prevent the forgetting of the base classes and avoid overfitting of the novel classes.

~\citet{qin2022continual} define relation learning in few-shot and incremental scenarios as continual few-shot relation learning and propose a method based on embedding space regularization and data augmentation to solve this problem.
~\citet{wangFewShotClassIncrementalLearning2022} use the generation-replay method to solve FSCIL for named entity recognition, which generates synthetic data of old entity classes for distillation.
~\citet{qin2021lfpt5} propose a unified framework for lifelong few-shot language learning, LFPT5, based on prompt tuning of T5. LFPT5 performs well on three different tasks: sequence labeling, text classification, and text generation, and is suitable for real-world applications.

In addition, FSIL has also been applied to the fusion field of images and NLP. For example, in the label-to-image translation field, which uses deep learning algorithms to learn the mapping relationship from semantic space to image space.~\citet{chenFewShotIncrementalLearning2022} propose a FSIL method for label-to-image translation, which solves this task with semantically-adaptive filters and normalization.

\subsection{Few-shot incremental learning in graph}
Recent studies have applied FSCIL to graphs~\citep{tan2022graph, lu2022geometer}. To maintain consistency with existing literature, we refer to this as graph Few-shot class incremental learning (GFSCIL). One of the pioneering studies in this field is the HAG-Meta method proposed by~\citet{tan2022graph}, which incorporates the previously mentioned Prospective Learning concept. HAG-Meta is based on the graph pseudo incremental learning paradigm and enables the model to learn new classes incrementally by cyclically adopting them from the base classes. Furthermore, it addresses class imbalance problems using hierarchical-attention-based modules.
~\citet{lu2022geometer} proposed Geometer to tackle GFSCIL problems. Geometer predicts the label of a node by identifying the nearest class prototype in the metric space and adjusts the attention-based prototypes by observing the geometric proximity, uniformity, and separability of novel classes. To mitigate catastrophic forgetting and unbalanced labeling issues, teacher-student knowledge distillation and biased sampling are also introduced. However, both of these methods are unable to handle dynamic graph structures.

\section{Future works} \label{sec-7}
% In this section, we discuss the possible future directions of the class-incremental learning field.

In this section, we discuss three key directions for the further development of FSCIL, namely, (i) theories, (ii) FSCIL settings and (iii) applications.

\subsection{Theories}
In order to further advance the field of FSCIL, there are several key areas that require attention in future research. Firstly, researchers should aim to enhance the efficiency of the algorithm by considering both performance and complexity. While many studies have solely focused on improving performance, it is important to also take into account the resource requirements of these methods.
Secondly, it is crucial to improve testing standards to more accurately evaluate performance across multiple tasks and on the base dataset. Although the average accuracy metric is widely used, it fails to account for the issue of imbalanced base classes and novel classes data. Additionally, the  performance dropping rate solely focuses on accuracy of the base and final tasks, without considering the accuracy of intermediate processes. In comparison, relative performance dropping rate~\citep{tan2022graph} and harmonic accuracy~\citep{pengFewShotClassIncrementalLearning2022} offer more comprehensive means of measuring model performance. Thirdly, as the ViT~\citep{dosovitskiy2021an} continues to gain importance, it may be worthwhile to explore its potential for use in FSCIL, as exemplified in~\citet{zhouFewShotClassIncrementalLearning2022}. By addressing these key areas, future research can build upon the current state-of-the-art and continue to advance this important area of machine learning.

\subsection{FSCIL settings}
The current experimental guidelines for FSCIL largely follow the setting proposed in~\citet{taoFewShotClassIncrementalLearning2020a}, which assumes a fixed number of new classes and samples per class in each incremental phase. However, this setting is difficult to meet in real-world applications. To better address this issue,~\citet{Ahmad_2022_CVPR} extended FSCIL to be variable, where in each incremental session, a learning agent can expect up to N ways and up to K shots. Additionally,~\citet{kalla2022s3c} proposed a more general setting, where novel classes have different numbers of samples, known as FSCIL-imbalanced, and the number of base classes is not abundant, known as FSCIL-less base. Exploring approaches closer to real-world applications, such as how to handle variable numbers of new classes and shots in different sessions, has practical significance. It is also worth investigating the fusion of FSL with Task-IL and Domain-IL, which are promising research directions.

\subsection{Applications}
The application of FSCIL in various interdisciplinary fields is a promising avenue for exploration in the future. For instance, recent research has introduced FSIL into the field of audio~\citep{wang2021few}, dynamic few-shot learning for multi-label audio classification~\citep{gidaris2018dynamic}, automatic radar modulation recognition~\citep{luo2022new}, intrusion detection~\citep{wang202intrusiondetection}, and medical time-series classification~\citep{sun2023few}. However, these methods are limited to single-scene settings, thus lacking scalability. Therefore, establishing a unified theoretical framework that is applicable to a wide range of scenarios is one of the future directions to address complex and multimodal tasks.

% 视觉语言模型：CLIP、CoOP、Co-CoOP
\section{Conclusion} \label{sec-8}
Few-shot class-incremental learning is a challenging yet crucial task. It reflects how humans learn in real-world scenarios where high-quality data is often limited and learning data is continually presented. In this paper, we have provided a comprehensive survey of existing FSCIL approaches and attempted to categorize them into five families, including traditional machine learning methods, meta learning-based methods, feature and feature space-based methods, replay-based methods, and dynamic network structure-based methods. 
Integrating these methodologies to balance performance, scalability, efficiency, and complexity may provide a direction for future research.
We have also discussed the performance of classic FSCIL methods and the applications of FSCIL in various fields of deep learning. However, FSCIL remains an underexplored area, and further research is required to explore its potential applications and theories.  Due to limitations of space, some theoretical derivations of the content were not extensively introduced. With the increasing demand for real-world AI applications, FSCIL research will continue to attract more attention and drive new innovations in the field of deep learning. 





\section*{Acknowledgements}
This work is supported by the National Natural Science
Foundation of China (Grant No. 62373343) and the Beijing Natural Science Foundation (No. L233036).
\bibliography{refs.bib}
\end{document}